\documentclass[11pt,leqno]{article}
\usepackage{authblk}
\usepackage[T1]{fontenc}
\usepackage{amsmath}%
\usepackage{amsthm}
\usepackage{amsxtra}%
\usepackage{amsfonts}%
\usepackage{amssymb}%
\usepackage[margin=1.25in]{geometry}
\usepackage{color,hyperref}
\usepackage{graphicx}
\usepackage{caption,subcaption}
\usepackage[scientific-notation=true]{siunitx}
\usepackage{cite}
\hypersetup{colorlinks,breaklinks,
             linkcolor=blue,urlcolor=blue,
             anchorcolor=blue,citecolor=blue}
\usepackage{color}
\usepackage{array}
\usepackage{booktabs}
\usepackage{enumerate}
\usepackage{dsfont}
\usepackage{hhline}

\DeclareMathOperator*{\supp}{supp}

\renewcommand{\bar}[1]{{\overline{#1}}}

\newcommand{\R}{\mathbb{R}}

\renewcommand{\phi}{\varphi}

\renewcommand{\epsilon}{\varepsilon}

\newcommand{\bx}{\mathbf{x}}

\newcommand{\by}{\mathbf{y}}
\newcommand{\bu}{\mathbf{u}}
\newcommand{\bz}{\mathbf{z}}
\newcommand{\sig}{\sigma}

\renewcommand{\th}{\theta}
\newcommand{\tP}{\Tilde{P}}
\newcommand{\tQ}{\Tilde{Q}}
\newcommand{\bw}{\mathbf{w}}
\newcommand{\lan}{\langle}
\newcommand{\ran}{\rangle}

\newcommand{\sumj}{\sum_{j=1}^N}
\renewcommand{\a}{\alpha}

\def\XXint#1#2#3{{\setbox0=\hbox{$#1{#2#3}{\int}$ }
\vcenter{\hbox{$#2#3$ }}\kern-.6\wd0}}

\newtheorem{theorem}{Theorem}

\newtheorem{proposition}[theorem]{Proposition}
\theoremstyle{definition}
\newtheorem{remark}[theorem]{Remark}

\numberwithin{equation}{section}
\numberwithin{theorem}{section}

\title{Attraction-Repulsion Swarming: A Generalized Framework of t-SNE via Force Normalization and Tunable Interactions\thanks{ {\bf Funding:} Both authors acknowledge the support of NSF-CCF:2212318. JC was also supported by the Alfred P. Sloan Foundation, and an Albert and Dorothy Marden Professorship. {\bf Source Code:} \url{https://github.com/jwcalder/AttractionRepulsionSwarming-tSNE}}}
\author{Jingcheng Lu, Jeff Calder}
\affil{School of Mathematics, University of Minnesota\thanks{{\bf Emails:} \textit{lujingcheng666@gmail.com, jwcalder@umn.edu}}}

\begin{document} 

\maketitle

\begin{abstract}
We propose a new method for data visualization based on attraction-repulsion swarming (ARS) dynamics, which we call ARS visualization. ARS is a generalized framework that is based on viewing the t-distributed stochastic neighbor embedding (t-SNE) visualization technique as a swarm of interacting agents driven by attraction and repulsion. Motivated by recent developments in swarming, we modify the t-SNE dynamics to include a normalization by the \emph{total influence}, which results in better posed dynamics in which we can use a data size independent time step (of $h=1$) and a simple iteration, without the need for the array of optimization tricks employed in t-SNE. ARS also includes the ability to separately tune the attraction and repulsion kernels, which gives the user control over the tightness within clusters and the spacing between them in the visualization. 

In contrast with t-SNE, our proposed ARS data visualization method is not gradient descent on the Kullback-Leibler divergence, and can be viewed solely  as an interacting particle system driven by attraction and repulsion forces. We provide theoretical results illustrating how the choice of interaction kernel affects the dynamics, and experimental results to validate our method and compare to t-SNE on the MNIST and Cifar-10 data sets. 
\end{abstract}

\section{Introduction}

Visualization of high-dimensional data is an important field of research in data analysis, as it helps to develop an intuitive understanding of data and formulate statistical hypotheses on complex data sets. In recent years, the t-distributed stochastic neighbor embedding (t-SNE) \cite{van2008visualizing} method for data visualizations has become one of the most popular techniques for dimensionality reduction. It has been widely applied in many fields, e.g., computer security \cite{hamid2020t,xue2020classification,yi2021anomaly,zoghi2021unsw,olszewski2024dimensionality}, cancer research \cite{abdelmoula2016data,dolezal2018diagnostic,gang2018dimensionality,mandel2019expression,mandel2020sequential,sharma2021stochastic,bocker2022toward,zhou2022classification}, bio-informatics \cite{platzer2013visualization,taskesen20162d,li2017application,kobak2019,linderman2019fast,cieslak2020t}, and natural language processing \cite{wang2018comparison,sanders2021unmasking,shetty2021automated,miyamoto2022natural,sharma2023optimization}.   The method has also been studied mathematically, with a focus on scaling limits, understanding its ability to uncover clustering structure in data, and the role played by early exaggeration, an optimization trick we discuss in more detail below  \cite{steinerberger2022t,linderman2019clustering,linderman2022dimensionality,murray2024large}.

The main idea behind the t-SNE visualization method is to construct a low dimensional embedding of a data set --- usually into $\R^2$ or $\R^3$ --- that preserves \emph{local} structure in the data, while allowing distortions at larger scales. This is accomplished by constructing a localized similarity weight matrix over the high dimensional data set, a heavy-tailed similarity matrix over the embedded data, normalizing both to be probability distributions, and then minimizing the Kullback-Leibler divergence between them. This acts to ensure that the localized structure of the embedding is faithful to the original high dimensional data.

The t-SNE embedding is computed through gradient-based optimization of the Kullback-Leibler divergence. However, it is well-known that plain gradient descent on the Kullback-Leibler divergence is extremely slow to converge and can often give poor visualization results. In order to achieve good results, it is necessary to use a collection of optimization tricks, including early exaggeration --- amplifying the attraction forces in the early stages --- gradient clipping, momentum, as well as scaling the time step with the data size. Good rules of thumb are available for how to tune the corresponding parameters, but their roles in producing good visualizations are only partially understood \cite{linderman2022dimensionality}.

Viewing each embedded data point as a particle in $\R^2$ or $\R^3$, the gradient descent dynamics of t-SNE break down into attraction and repulsion forces driving the embedded points --- the attraction terms encourge the embedding to preserve local structure, while the repulsion terms prevent collapse to a trival solution. This viewpoint was given in the seminal work \cite{van2008visualizing}, and has been expounded upon recently in \cite{linderman2022dimensionality}. In fact, Linderman and Steinerberger  \cite{linderman2022dimensionality} go so far as to pose the question of how to construct a dynamical system of attraction-repulsion type whose evolution is well-suited for data visualization.

In this paper, we follow this point of view, and propose a data visualization method we call Attraction-Repulsion Swarming (ARS) visualization. Instead of minimizing the Kullback-Leibler divergence, ARS visualization is the steady-state of attraction-repulsion particle dynamics, or swarming. Motivated by recent work in swarming-based optimization approaches, we scale the dynamics by the \emph{total influence}, which results in well-posed dynamics that converge to steady state extremely quickly, without the need to employ \emph{any} optimization tricks. Furthermore, we can use the same time step of $h=1$, \emph{independent} of the size of the data. 

The ARS dynamics also include the ability to separately tune the strengths of the attraction and repulsion kernels, which allows the user to control the tightness of the clusters and the separation between them, in the visualization. We find that using stronger attraction forces and weaker repulsion leads to better visualizations. Other works have modified the tails of the similarity kernel in t-SNE \cite{yang2009heavy,kobak2019heavy}, but t-SNE is unable to modify the attraction and repulsion forces separately 

\begin{figure}[t!]
\centering
\begin{subfigure}{0.48\textwidth}
\includegraphics[width=\textwidth]{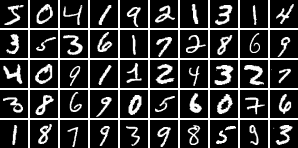}    
\subcaption{MNIST}
\end{subfigure}
\hfill
\begin{subfigure}{0.48\textwidth}
\includegraphics[width=\textwidth]{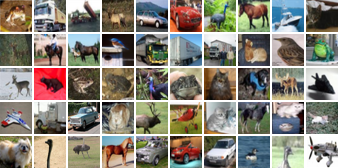}    
\subcaption{Cifar-10}
\end{subfigure}
\caption{Examples of MNIST and Cifar-10 images.}
\label{fig:examples}
\end{figure}

\begin{figure}[t!]
\centering
\begin{subfigure}{0.32\textwidth}
\includegraphics[width=\textwidth]{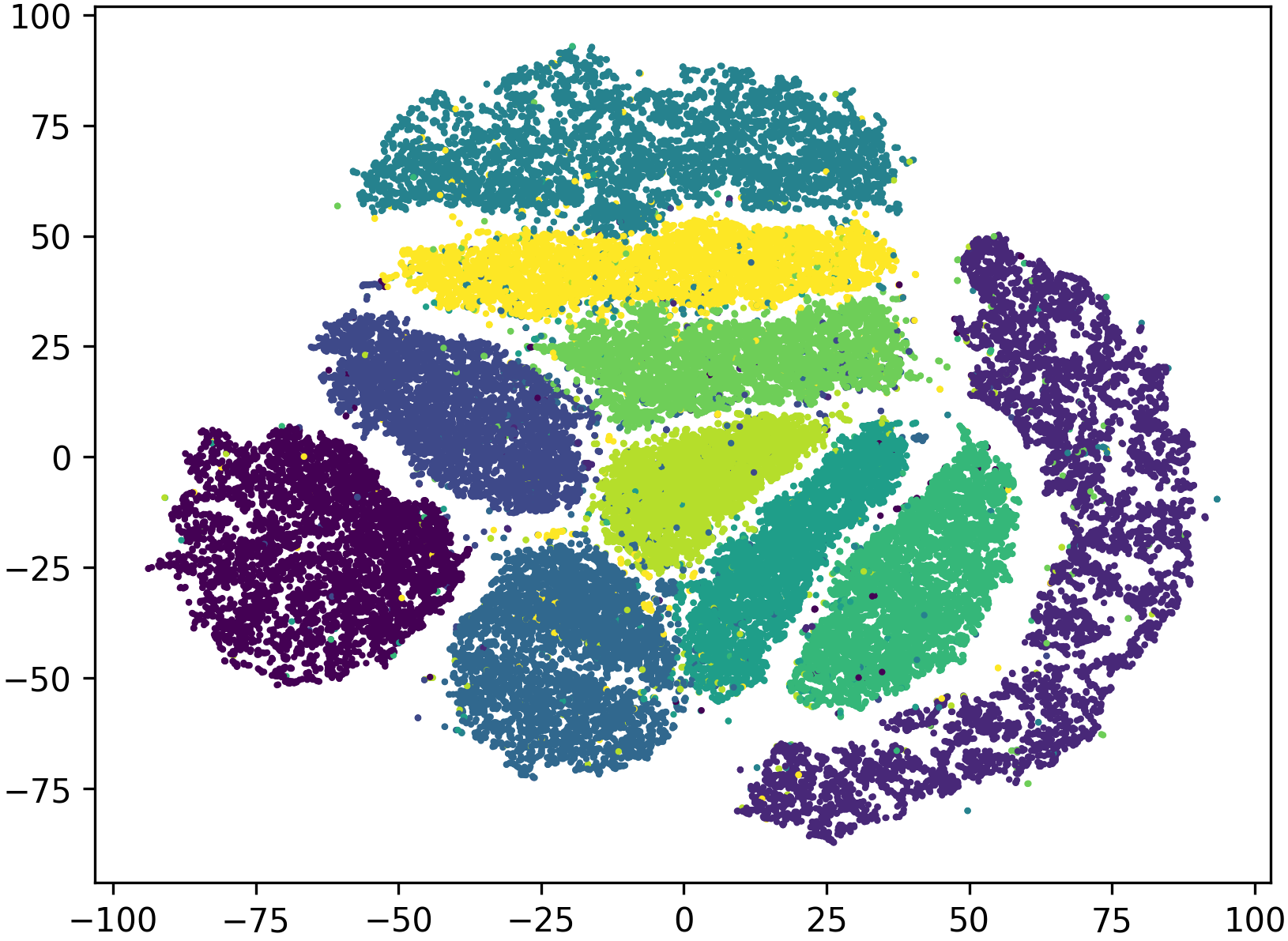}    
\subcaption{MNIST: ARS$_{2,2}$}
\end{subfigure}
\begin{subfigure}{0.32\textwidth}
\includegraphics[width=\textwidth]{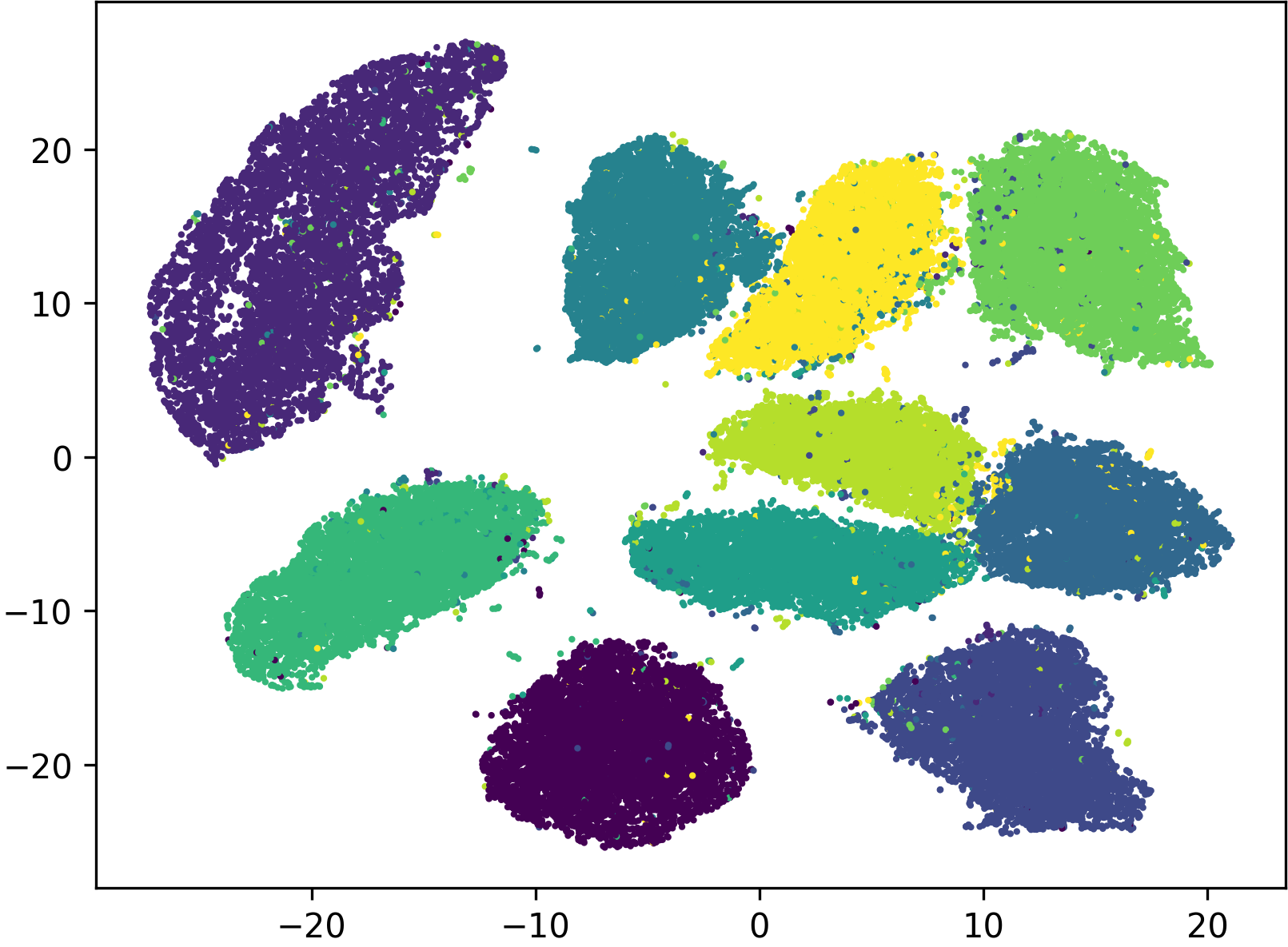}    
\subcaption{MNIST: ARS$_{2,3}$}
\end{subfigure}
\begin{subfigure}{0.32\textwidth}
\includegraphics[width=\textwidth]{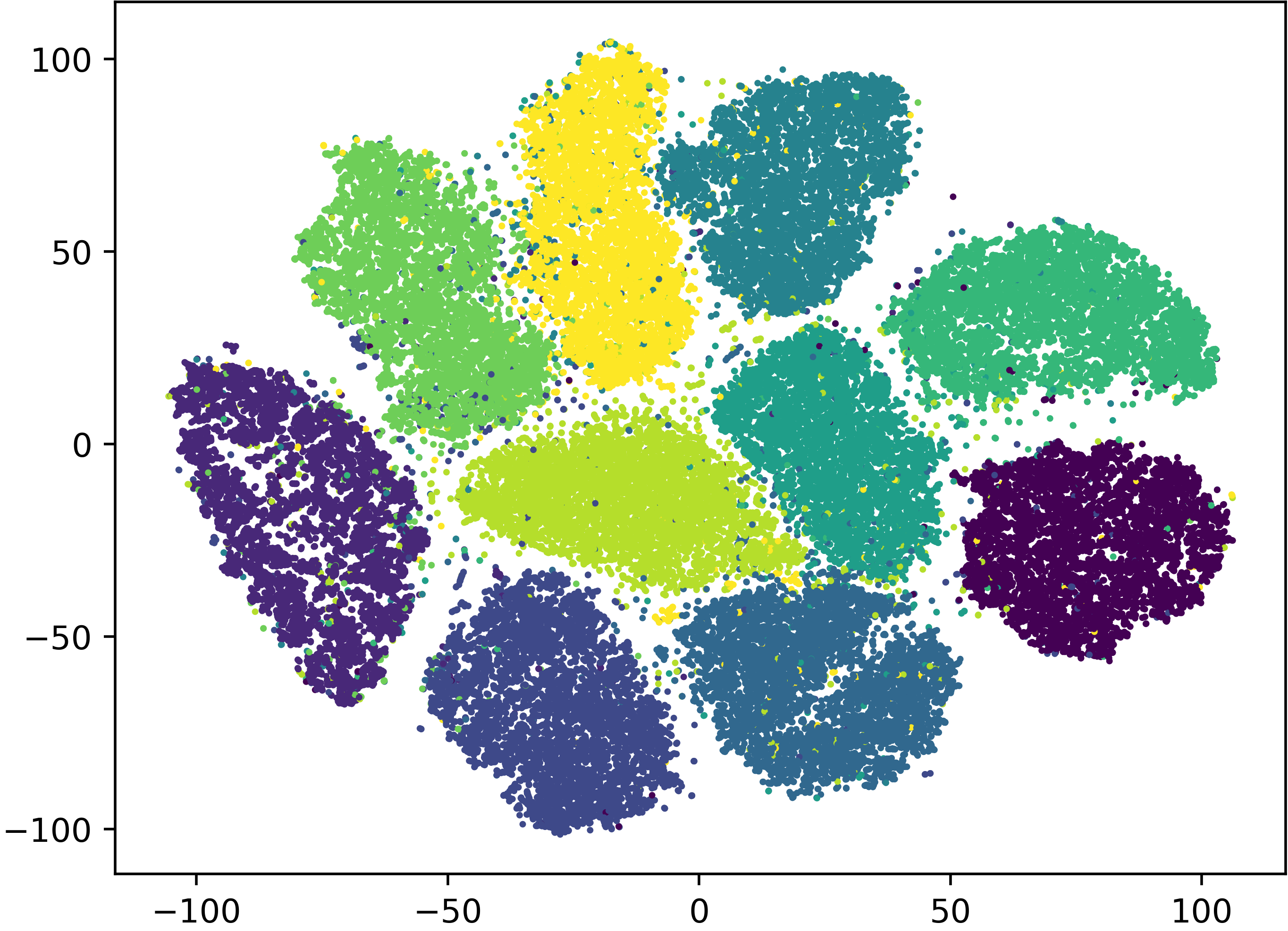}    
\subcaption{MNIST: t-SNE}
\end{subfigure}\\
\begin{subfigure}{0.32\textwidth}
\includegraphics[width=\textwidth]{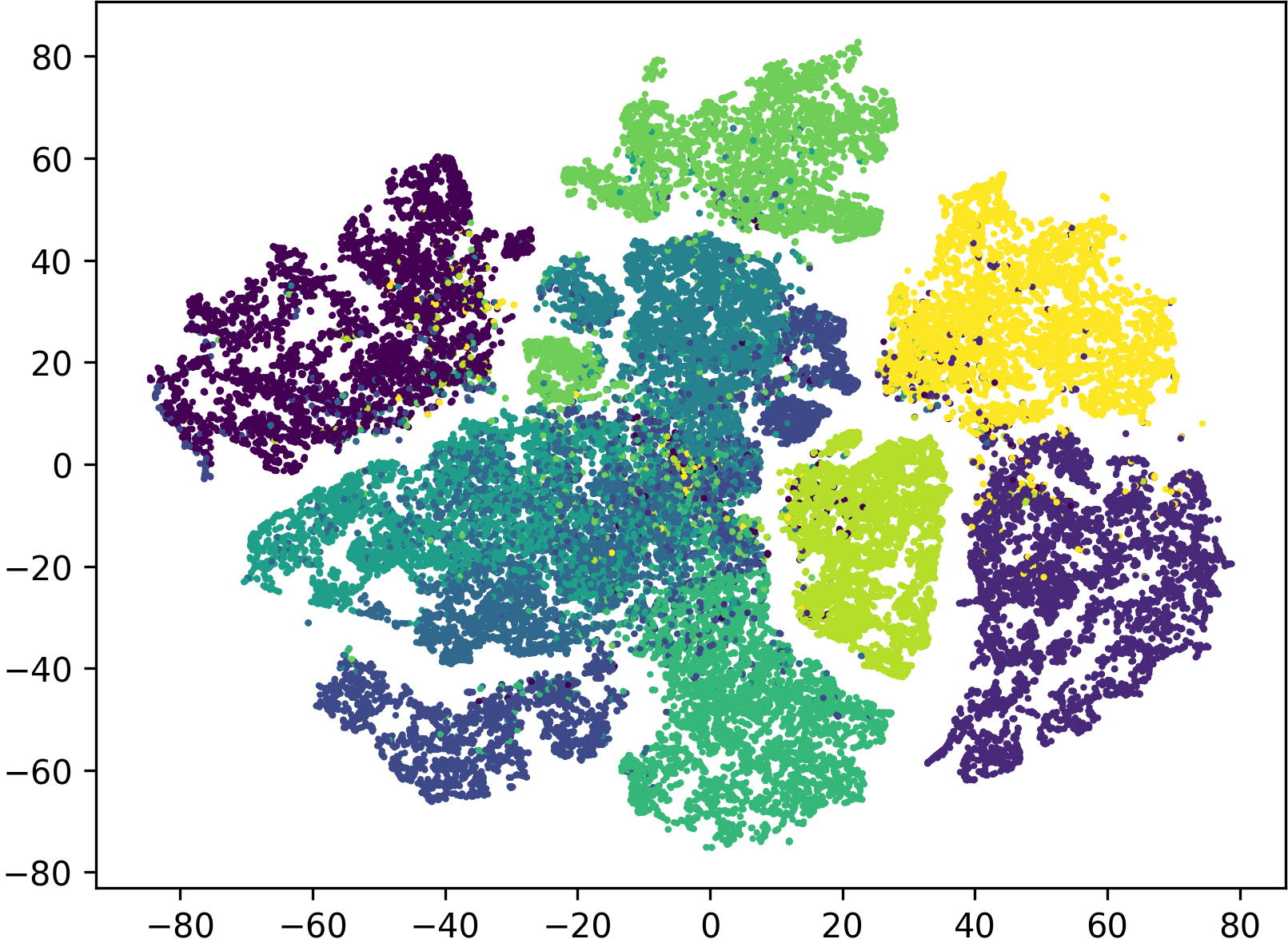}    
\subcaption{Cifar-10: ARS$_{2,2}$}
\end{subfigure}
\begin{subfigure}{0.32\textwidth}
\includegraphics[width=\textwidth]{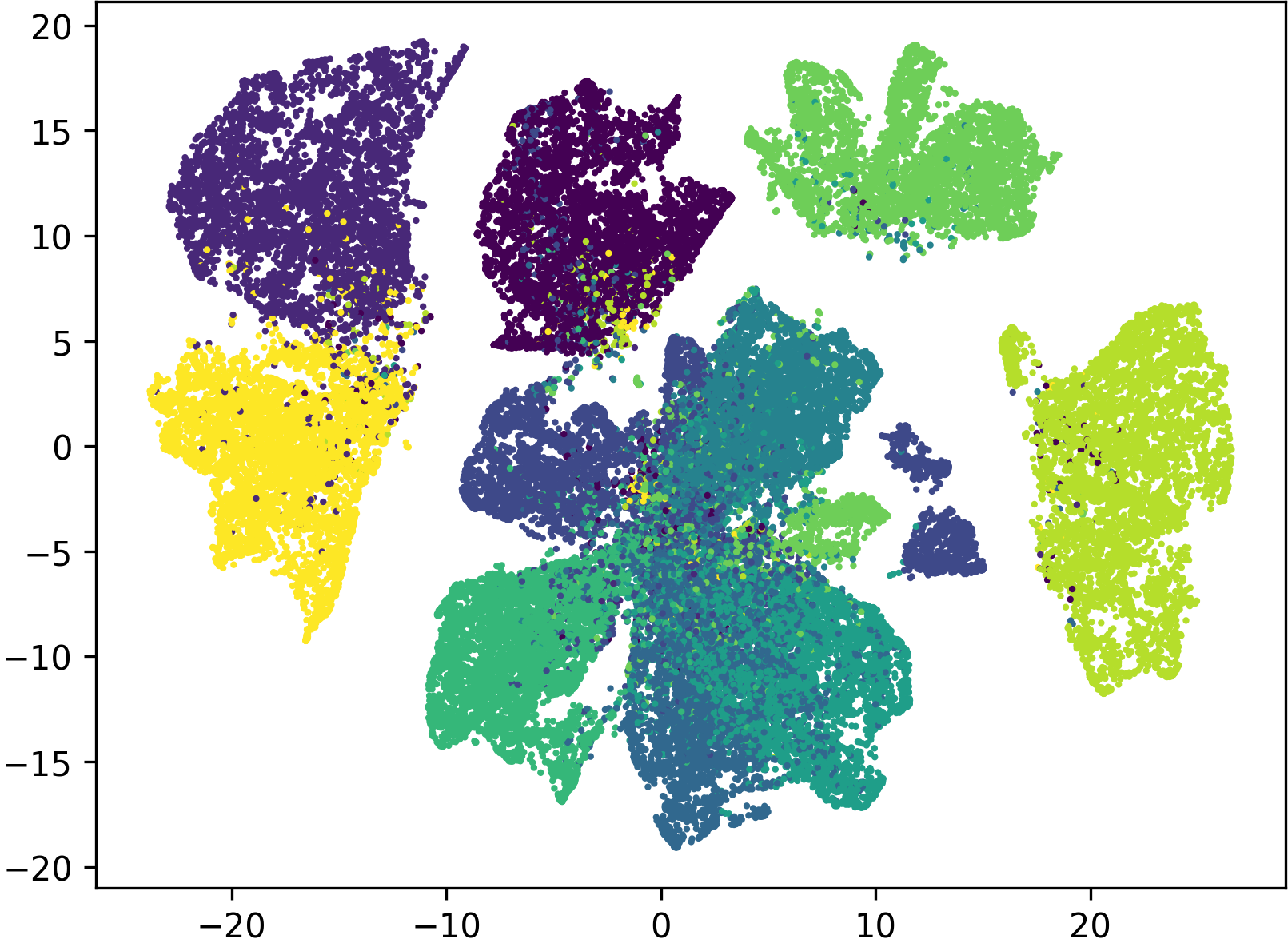}    
\subcaption{Cifar-10: ARS$_{2,3}$}
\end{subfigure}
\begin{subfigure}{0.32\textwidth}
\includegraphics[width=\textwidth]{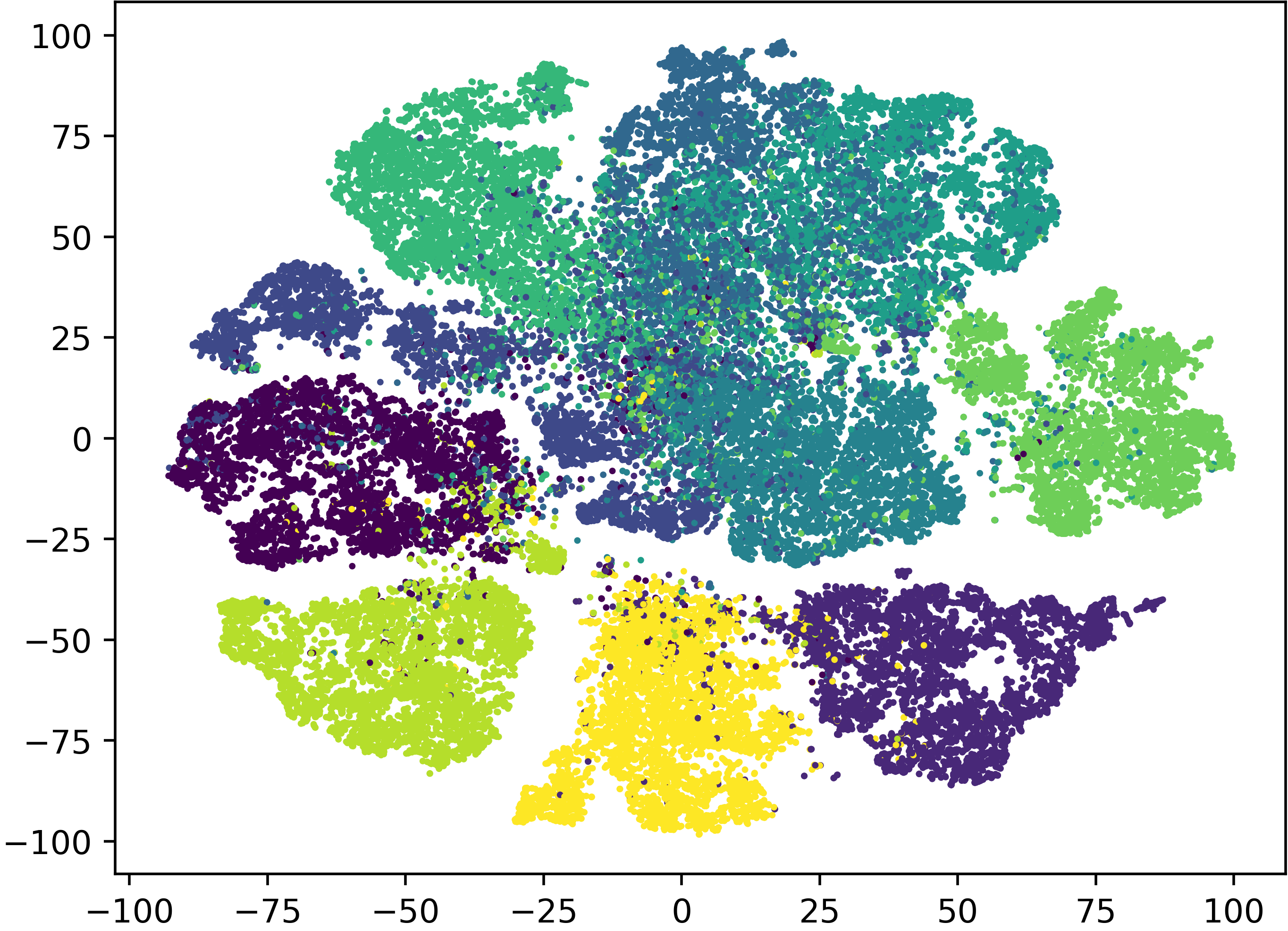}    
\subcaption{Cifar-10: t-SNE}
\end{subfigure}
\caption{Comparison of ARS and t-SNE on MNIST and Cifar-10.}
\label{fig:ars_tsne}
\end{figure}

To illustrate ARS visualization in comparison with t-SNE, we show here the results of both methods applied to visualize the MNIST \cite{lecun1998gradient} and Cifar-10 \cite{krizhevsky2009learning} image data sets. Sample images from each data set are shown in Figure \ref{fig:examples}. For MNIST we represented the images with their raw pixel values, while for Cifar-10 we used the SimCLR contrastive learning embedding \cite{chen2020simple}. Figure \ref{fig:ars_tsne} shows visualizations of the full MNIST and Cifar-10 data sets using t-SNE and two versions of ARS visualization.\footnote{We used the sklearn version of t-SNE with default parameters and a perplexity of $30$. The MNIST data set has $70000$ images, while Cifar-10 has $60000$.} The ARS$_{2,2}$ method has balanced attraction and repulsion forces, and is most similar to t-SNE, while ARS$_{2,3}$ has stronger attraction forces and weaker repulsion, leading to tighter clusters with better separation between clusters --- more details are given in Section \ref{sec:ars}. We generally recommend the use of ARS$_{2,3}$, and we describe these numerical results in more detail in Section \ref{sec: ARS-BH}.

The ability of ARS to give comparable results to t-SNE strongly suggests that the attraction-repulsion dynamics in t-SNE are the main mechanisms behind producing good visualizations, and the Kullback-Leibler divergence is not an essential part of the method. Furthermore, our numerical results suggest that minimizing the Kullback-Leibler divergence is exactly what leads to slow convergence of t-SNE and the need to devise complicated optimization tricks to speed up convergence and improve results. By switching to the dynamical attraction-repulsion perspective, we are able to produce dynamics that converge quickly without the need to employ advanced optimization tricks. 

The rest of this paper is organized as follows. In Section \ref{sec:ars} we introduce our ARS visualization method. In Section \ref{sec:datasize} we show how the time step in ARS is independent of the data set size. In Section \ref{sec:tune} we explore the role of tunable interaction kernels in ARS. In Section \ref{sec:meanfield} we give some preliminary arguments showing that ARS may be studied rigorously through a mean-field limit. Finally, in Section \label{sec: ARS-BH} we present numerical results comparing ARS to t-SNE on the large scale data sets MNIST and Cifar-10.

\section{Attraction-Repulsion Swarming Visualization}\label{sec:ars}

In this section, we overview the t-SNE visualization technique and then introduce our proposed Attraction-Repulsion Swarming (ARS) visualization. 

\subsection{A Brief Review of t-SNE}

The purpose of t-SNE is to project a set of $D$-dimensional data points, $\{\bx_i\}^N_{i=1}$, $\bx_i\in \R^D$, into two or three-dimensional space while maintaining the essential \emph{local} similarity features in the data. Specifically, the similarity of data point $\bx_j$ to data point $\bx_i$ is measured by the conditional probability,
\begin{equation}\label{eq:tsne pcond}
p_{j|i} = \frac{\exp(-|\bx_i-\bx_j|^2/2\sig_i^2)}{\sum_{k\neq i} \exp(-|\bx_i-\bx_k|^2/2\sig_i^2)}, \quad i\neq j, \quad p_{i|i} = 0,
\end{equation}
where $\sig_i$ is the bandwidth of the Gaussian kernel centered at $\bx_i$. The perplexity of $\bx_i$, which measures the effective number of neighbors, is defined as
\[
Perp(P_i) = 2^{H(P_i)}, \quad H(P_i) = -\sum_{j\neq i} p_{j|i}\log_2(p_{j|i}).
\]
Then bandwidths $\sig_i$ are chosen so that the perplexity of each data point equals a specified value, typically between 5 and 50. Once the conditional probabilities, $p_{j|i}$, are obtained, the global similarity information is encoded into the symmetrized probability distribution, $P = (P_{ij})^N_{i,j=1}$, with
\begin{equation}
     P_{ij} = \frac{p_{i|j}+p_{j|i}}{2N}.
\end{equation}
It is easy to verify that $\displaystyle \sum^N_{i,j=1}P_{ij}=1$, so the matrix $P$ can be viewed as a probability distribution. The perplexity graph construction defined above shares similarities with a $k$-nearest neighbor graph, where the desired perplexity plays the role of the number $k$ of neighbors.

Associated to the high-dimensional data points $\{\bx_i\}^N_{i=1}$, let $\{\by_i\}^N_{i=1}$ denote the low dimensional projections $\by_i\in \R^d$, where $d=2$ or $d=3$. In t-SNE, the similarity of $\by_i$ and $\by_j$ is measured through the t-distribution
\begin{equation}
 Q_{ij} = \left\{\begin{array}{ll}
    \displaystyle\frac{1}{z(1+|\by_i-\by_j|^2)}, \quad & i\neq j\\ \\
    0 & i = j
    \end{array}\right., \ \ \text{where} \ \ \displaystyle z = \sum_k\sum_{l\neq k}(1+|\by_k-\by_l|^2)^{-1}.
\end{equation}
The normalization coefficient $z$ is chosen to ensure that $\sum^N_{i,j=1}Q_{ij} = 1$, so that the matrix $Q$, like $P$, is a probability distribution. The t-SNE algorithm optimizes the positions of the embedded points $\by_i$ by minimizing the Kullback-Leibler divergence
\begin{equation}\label{eq:KL divergence}
    E(\by_1,\by_2,\cdots,\by_N) = KL(P||Q) = \sum^{N}_{i,j = 1}P_{ij}\log(\frac{P_{ij}}{Q_{ij}}).
\end{equation}
This cost function provides a measure of how well the distribution $Q_{ij}$ models $P_{ij}$ in low-dimensional space. The particular case of $P_{ij} = Q_{ij}$ yields $KL(P||Q) = KL(P||P) = 0$. 

The t-SNE cost function is minimized with gradient-based optimization techniques. In this case, the gradient of the Kullback-Leibler divergence cost function \eqref{eq:KL divergence} is given by 
\[
\begin{split}
\frac{\partial E}{\partial \by_i} &= 4z\sum_{j=1}^N(P_{ij}-Q_{ij})Q_{ij}(\by_i-\by_j) 
\end{split}.
\]
The gradient descent iterations under step size (or learning rate) $h$ then read
\begin{equation}\label{eq: KL GD}
\begin{split}
    \by_i^{n+1} &= \by^{n}_i+4zh\sum^N_{j=1}(P_{ij}-Q_{ij})Q_{ij}(\by^n_j-\by^n_i)\\
    & = \by_i^n+4zh\sum_{j=1}^NP_{ij}Q_{ij}(\by^n_j-\by^n_i)-4zh\sum_{j=1}^N Q^2_{ij}(\by^n_j-\by^n_i).
\end{split}
\end{equation}
This dynamics can also be regarded as the forward Euler discretization of the following gradient flow under time step $4h$,
\begin{equation}\label{eq: KL grad flow}
\begin{split}
    \frac{d}{dt}\by_i(t) &= z\sum_{j=1}^NP_{ij}Q_{ij}(\by_j-\by_i)-z\sum_{j=1}^N Q^2_{ij}(\by_j-\by_i)\\
    & = \sum_{j=1}^NP_{ij}\psi(|\by_i-\by_j|)(\by_j-\by_i)-\sum_{j=1}^N Q_{ij}\psi(|\by_i-\by_j|)(\by_j-\by_i),
\end{split}
\end{equation}
where the interaction kernel $\psi$ is given by
\[
\psi(|\by_i-\by_j|) = zQ_{ij} = \frac{1}{1+|\by_i-\by_j|^2}.
\]
The dynamics \eqref{eq: KL grad flow} is comprised of two parts: the first term on the right side
\[
F_{attr,i} = \sum^N_{j=1} P_{ij}\psi(|\by_i-\by_j|)(\by_j-\by_i) 
\]
is an attraction term which pulls similar data points together. The second term
\[
 F_{rep,i} =  -\sum^N_{j=1} Q_{ij}\psi(|\by_i-\by_j|)(\by_j-\by_i)
\]
serves a repulsion term that attempts to separate nearby points. The kernel $\psi(\cdot)$ is radially decreasing, indicating that the intensity of inter-agent interactions is inversely dependent on their distance. The prefactors, $P_{ij}$ and $Q_{ij}$, further modify the amplitudes of attraction and repulsive forces based on the similarity and the distance between data points.

\subsection{Motivation for ARS}

The gradient descent iterations \eqref{eq: KL GD} often suffer from slow convergence and poor visualization results when applied to large scale complicated data sets. To address this, a classical approach \cite{van2008visualizing} is to introduce \emph{early exaggeration}, which corresponds to the dynamics
\begin{equation}\label{eq:early exaggeration}
 \by_i^{n+1}  = \by^n_i+4zh\a\sumj P_{ij}Q_{ij}(\by^n_j-\by^n_i)-4zh\sumj  Q^2_{ij}(\by^n_j-\by^n_i), \quad \a >1.
\end{equation}
The factor $\a >1$ amplifies the attraction effect on the global configuration, thereby gathering similar data points tightly into clusters. After a desired number of early exaggeration steps \eqref{eq:early exaggeration}, often around $200$, we return to the gradient flow dynamics \eqref{eq: KL GD} and the repulsive forces between dissimilar agents are restored, causing the congested crowd to spread out into separated clusters more evenly.  

In addition to early exaggeration, it is necessary to use a very large learning rate $h$, often between 10 to 1000, to accelerate convergence and avoid getting stuck at a poor solution. Indeed, notice that the average scales of $P_{ij}$, $Q_{ij}$ are $O(1/N^2)$, hence the total force scales with $O(1/N)$, where $N$ is the number of data points. To compensate for the $1/N$ shrinkage of the force magnitude for large data size $N\gg 1$, the appropriate learning rate should scale with $O(N)$.  However, an excessively large learning rate can also result in the risk of instability or poor results --- if the learning rate is taken too large, the data may end up looking like a `ball', where each agent is approximately equidistant from its nearest neighbors. To stabilize the convergence under large learning rates, common implementations of t-SNE in the existing literature set a relatively large initial learning rate (e.g., $N/\a$ proposed in \cite{belkina2019,kobak2019}), followed by learning rate adaptation techniques to ensure stability. 

One of the most popular adaptation schemes is the delta-bar-delta rule proposed by Jacob \cite{jacobs1988increased}. It has been applied in many existing studies of t-SNE e.g., \cite{van2008visualizing,van2014accelerating, belkina2019,kobak2019}. The key idea of the scheme is to gradually increase the learning rate if the gradient direction remains stable. If the gradient changes sign at consecutive iterations, which indicates oscillation due to an excessively large learning rate, then the learning rate is decreased. More specifically, let $w^n$ be a single weight of the cost function $E(\bw^n)$ at the $n-$th iteration, the corresponding learning rate, $h^n$, is updated with the rule
\begin{equation}\label{eq:jacobs adaptation}
h^{n+1} = \left\{\begin{array}{ll}
    h^n+\kappa &, \bar{\delta}^{n-1}\cdot \delta^n>0 \\
    (1-\phi)h^n &, \bar{\delta}^{n-1}\cdot \delta^n<0
\end{array}
\right., \quad \kappa>0, \quad 0<\phi<1,
\end{equation}
where 
\[
\delta^n = \frac{\partial E(\bw^n)}{\partial w^n}, \ \ \text{and} \ \ \bar{\delta}^n = (1-\th)\delta^n+\th\bar{\delta}^{n-1}, \quad 0<\theta<1.
\]
Employing the adaptive learning rate with scheme \eqref{eq:jacobs adaptation} introduces $O(N)$ computational cost, which is insignificant compared to the overall cost of the t-SNE algorithm. Nevertheless, the appropriate choice of parameters $\kappa$, $\phi$, and $\theta$ can be experimental, which may cause inconvenience to unfamiliar users when applied to general problems.

\subsection{ARS Visualization}

To speed up the clustering of data without resorting early exaggeration, time-step adaptation, momentum, and other advanced optimization techniques, we propose a new \emph{Attraction-Repulsion Swarming} ($ARS_{\th_1,\th_2}$) method. The algorithm is obtained by discretizing the following continuous-time dynamics,
\begin{subequations}\label{eq: ARS}
\begin{equation} \label{eq:ARS dynamics}
\begin{split}    
   & \frac{d}{dt}\by_i(t) = \sumj \tP_{ij}\psi_1(|\by_i-\by_j|)(\by_j-\by_i)-\sumj \tQ_{ij}(t)\psi_2(|\by_i-\by_j|)(\by_j-\by_i),\\
   & \tP_{ij} = \frac{P_{ij}}{\sum_{k}P_{ik}}, \quad \tQ_{ij} = \frac{Q_{ij}}{\sum_{k}Q_{ik}} = \frac{(1+|\by_i-\by_j|^2)^{-1}}{\sum_{k\neq i} (1+|\by_i-\by_k|^2)^{-1}},
\end{split}    
\end{equation}
where the attraction and repulsion kernels take the form
\begin{equation}\label{eq:ARS kernels}
    \psi_1(r) = (1+r^{\th_1})^{-1}, \quad \psi_2(r) = (1+r^{\th_2})^{-1}, \quad \th_1,\th_2>0.
\end{equation}
\end{subequations}
The particular case of $\tP_{ij} = P_{ij}$, $\tQ_{ij} = Q_{ij}$, and $\th_1 = \th_2 = 2$ amounts to the original t-SNE gradient flow \eqref{eq: KL grad flow}. The discrete dynamics of \eqref{eq: ARS} under the forward Euler time stepping is given by
\begin{equation}\label{eq:ARS discrete}
\by_i^{n+1} = \by_i^{n} + h\sumj\tP_{ij}\psi_1(|\by^n_i-\by^n_j|)(\by^n_j-\by^n_i)-h\sumj \tQ_{ij}\psi_2(|\by^n_i-\by^n_j|)(\by^n_j-\by^n_i)
\end{equation}
To achieve a better visualization, early exaggeration may be introduced to reinforce the formation and the separation of clusters,
\begin{equation}\label{eq:ARS early exaggeration}
\by_i^{n+1} = \by_i^{n} + \alpha h\sumj\tP_{ij}\psi_1(|\by^n_i-\by^n_j|)(\by^n_j-\by^n_i)-h\sumj \tQ_{ij}\psi_2(|\by^n_i-\by^n_j|)(\by^n_j-\by^n_i). 
\end{equation}
Note that \eqref{eq:ARS early exaggeration} is a direct extension of the t-SNE counterpart \eqref{eq:early exaggeration} under normalized forcing.

\medskip
We outline the key components of ARS below:

\paragraph{Force Normalization:} Compared with the original t-SNE algorithm, the forcing terms of ARS are further scaled by the \emph{total influence}, $\sum_j P_{ij}$ and $\sum_j Q_{ij}$, which is inspired by the flocking dynamics model proposed by Motsch and Tadmor \cite{motsch2011}. The new scaling emphasizes the \emph{`relative' (conditional)}  similarity and the relative distance in agent communications. The magnitudes of the force coefficients, $\tP_{ij}$ and $\tQ_{ij}$, are enlarged to $O(1/N)$, in contrast to $O(1/N^2)$ for $P_{ij}$ and $Q_{ij}$ in the original t-SNE algorithm. Thus, we obtain the modified scaling for attraction and repulsive forces,
\[
\begin{split}
& F_{attr,i} = \sum^N_{j=1}\tP_{ij}\psi_1(|\by_i-\by_j|)(\by_j-\by_i) \sim \frac{1}{N}\sum^N_{j=1} \psi_1(|\by_i-\by_j|)(\by_j-\by_i) \sim O(1), \\
& F_{rep,i} =  -\sum^N_{j=1}\tQ_{ij}\psi_2(|\by_i-\by_j|)(\by_j-\by_i) \sim -\frac{1}{N}\sum^N_{j=1} \psi_2(|\by_i-\by_j|)(\by_j-\by_i) \sim O(1).
\end{split}
\]
The dependence of the learning rate $h$ on the data size is subsequently removed, due to the $O(1)$ force magnitudes.

\paragraph{Tunable Interactions:} The new ARS dynamics allows distinct and adjustable communication protocols for attraction and repulsion, dictated by the parameters $\th_1$ and $\th_2$. With a smaller value of $\th_i$, the communication kernel $\psi_i$  admits a heavier tail, leading to stronger interactions at long range. In contrast, a greater value of $\th_i$ emphasizes short-range interaction and weakens the influence at far-field. While employing distinct interactions, $\th_1 \neq \th_2$, eliminates the equilibrium state $P_{ij} = Q_{ij}$, the quality of clustering can actually be improved, as will be shown through computations.

\begin{remark}
We point out that in the case of $\th_1 = \th_2 = 2$ (i.e., $\psi_1(r) = \psi_2(r) = (1+r^2)^{-1}$), the ARS$_{2,2}$ iterations can also be interpreted as a variant of the t-SNE iterations \eqref{eq: KL GD} with distinct step sizes for attraction and repulsion terms,   
\[\by^{n+1}_i = \by^{n}_i +  z h^{(1)}_i\sum^N_{j=1}P_{ij}Q_{ij}(\by^n_j-\by^n_i)- z h^{(2)}_i\sum^N_{j=1}Q^2_{ij}(\by^n_j-\by^n_i)\]
where 
\[
h^{(1)}_i = \frac{h}{\sum_j P_{ij}}, \quad h^{(2)}_i = \frac{h}{\sum_j Q_{ij}}.
\]
Compared with the classical step size adaptation protocols e.g., the scheme \eqref{eq:jacobs adaptation}, the learning rates $h^{(1)}_i$ and $h^{(2)}_i$ in the new ARS algorithm are scaled by the total influence, and are thus self-adjusting. This eliminates the need to experimentally set an array of sensitive optimization hyperparameters, which can strongly affect the actual computations, thereby making the ARS method more convenient to apply in general data visualization tasks.
\end{remark}

\section{Independence of Data Size}\label{sec:datasize}

A key aspect of ARS is the normalization of the forcing terms by \emph{total influence}, so that the learning rate is no longer required to scale with the data size. To verify this improvement, we compare the performance of t-SNE \eqref{eq: KL GD} and ARS \eqref{eq:ARS discrete} against the same 1000 MNIST images from the digits $0\sim 3$. Both methods employ the same initialization generated with the uniform distribution $\by^0_i \sim \mathcal{U}([0,1]^2)$. We used 100 steps of t-SNE early exaggeration \eqref{eq:early exaggeration} for both methods with $\a$ = 40.\footnote{Here we apply the t-SNE early exaggeration to ARS, instead of the normalized version \eqref{eq:ARS early exaggeration}, for the purpose of a fair comparison. For practical application of ARS, however, we still employ the normalized early exaggeration for convenient implementation. For relevant results, see Section \ref{sec: ARS-BH}.} We set the perplexity to 30.

\begin{figure}[t!]
    \centering
     \begin{subfigure}{0.32\textwidth}
    \includegraphics[width=\textwidth,clip=true,trim=45 25 55 35]{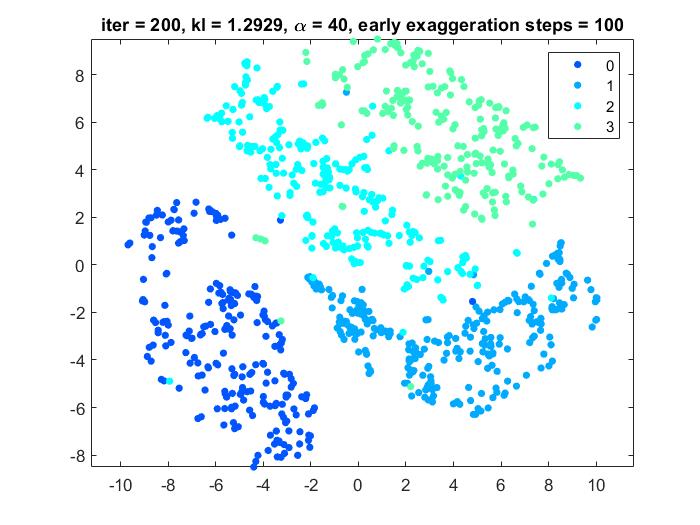}    
    \subcaption{ARS$_{2,2}$, $h = 1$, loss = 1.29}
    \end{subfigure}
     \begin{subfigure}{0.32\textwidth}
    \includegraphics[width=\textwidth,clip=true,trim=45 25 55 35]{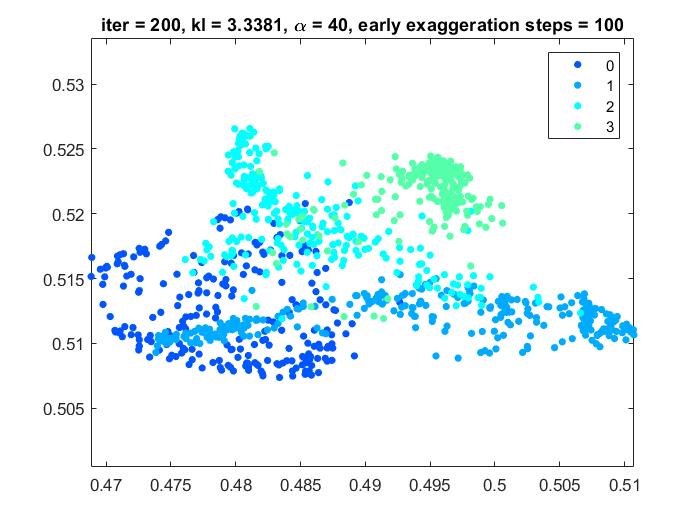}    
    \subcaption{t-SNE, $h = 1$, loss = 3.34}
    \end{subfigure}
    \begin{subfigure}{0.32\textwidth}
    \includegraphics[width=\textwidth,clip=true,trim=45 25 55 35]{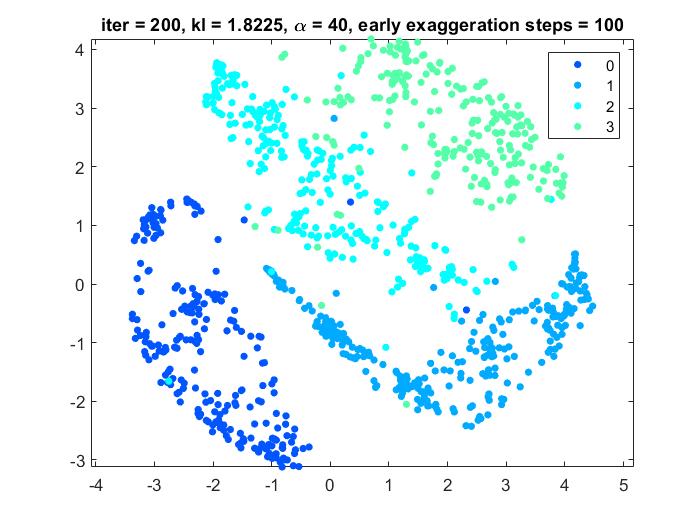}    
    \subcaption{t-SNE, $h = 70$, loss = 1.82}
    \end{subfigure}
    \\
  \begin{subfigure}{0.32\textwidth}
    \includegraphics[width=\textwidth,clip=true,trim=45 25 55 35]{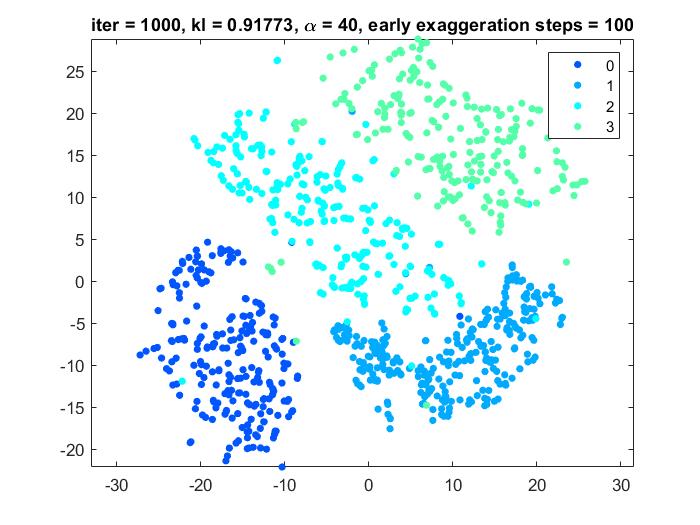}    
    \subcaption{ARS$_{2,2}$, $h = 1$, loss = 0.92}
    \end{subfigure}
     \begin{subfigure}{0.32\textwidth}
    \includegraphics[width=\textwidth,clip=true,trim=45 25 55 35]{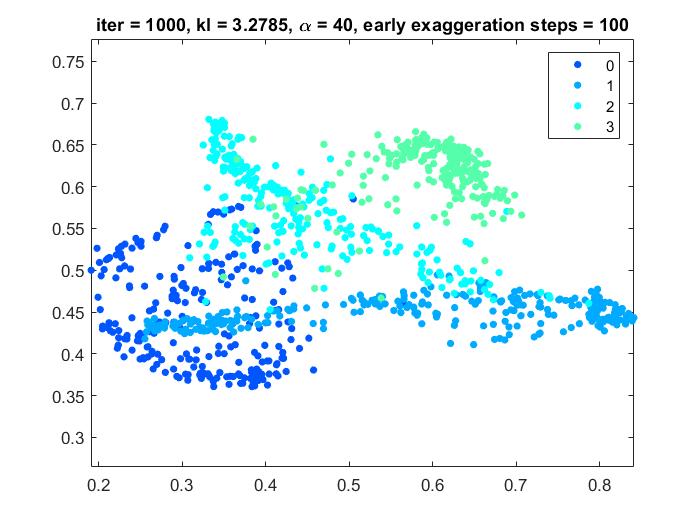}  
    \subcaption{t-SNE, $h = 1$, loss = 3.28}
    \end{subfigure}
     \begin{subfigure}{0.32\textwidth}
    \includegraphics[width=\textwidth,clip=true,trim=45 25 55 35]{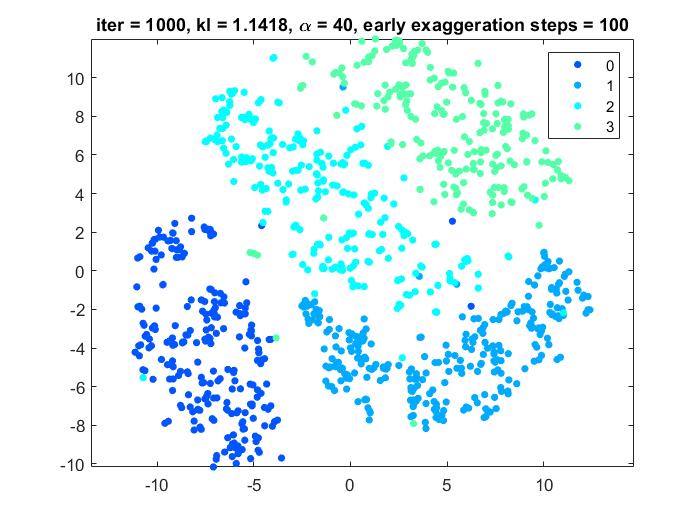}  
    \subcaption{t-SNE, $h = 70$, loss = 1.14}
    \end{subfigure}
    \caption{ARS$_{2,2}$ vs t-SNE applied to 1000 MNIST images of digits $0,1,2,3$, using $100$ steps of early exaggeration, $\a=40$. The top row corresponds to iteration $200$, while the bottom row is $1000$. For t-SNE, the time step $h$ is given for the gradient descent iterations, while $h=1$ is used during early exaggeration.}
    \label{fig: ARS vs GD N1000}
\end{figure}

\begin{figure}[t!]
    \centering
    \begin{subfigure}{0.32\textwidth}
    \includegraphics[width=\textwidth,clip=true,trim=55 25 55 35]{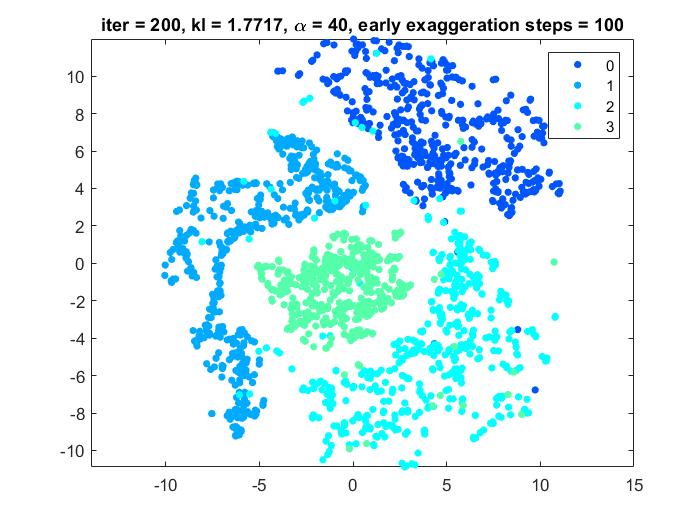}    
    \subcaption{ARS$_{2,2}$, $h = 1$, loss = 1.77}
    \end{subfigure}
    \begin{subfigure}{0.32\textwidth}
    \includegraphics[width=\textwidth,clip=true,trim=55 25 55 35]{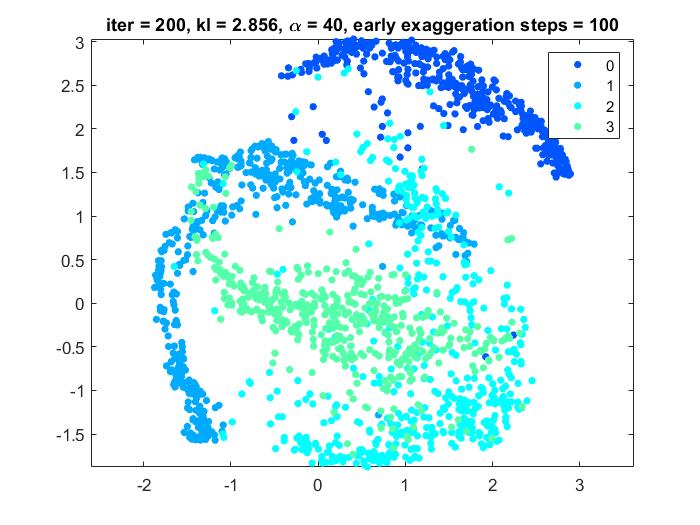}    
    \subcaption{t-SNE, $h = 70$, loss = 2.86}
    \end{subfigure}
     \begin{subfigure}{0.32\textwidth}
    \includegraphics[width=\textwidth,clip=true,trim=55 25 55 35]{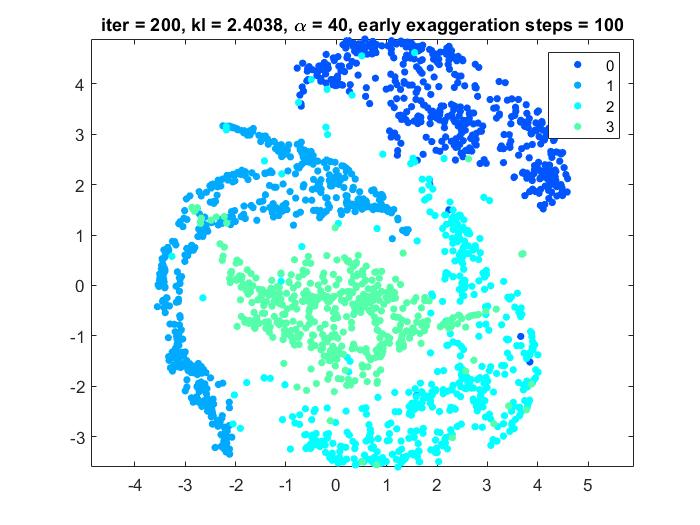}    
    \subcaption{t-SNE, $h = 150$, loss = 2.4}
    \end{subfigure}
    \\
    \begin{subfigure}{0.32\textwidth}
    \includegraphics[width=\textwidth,clip=true,trim=55 25 55 35]{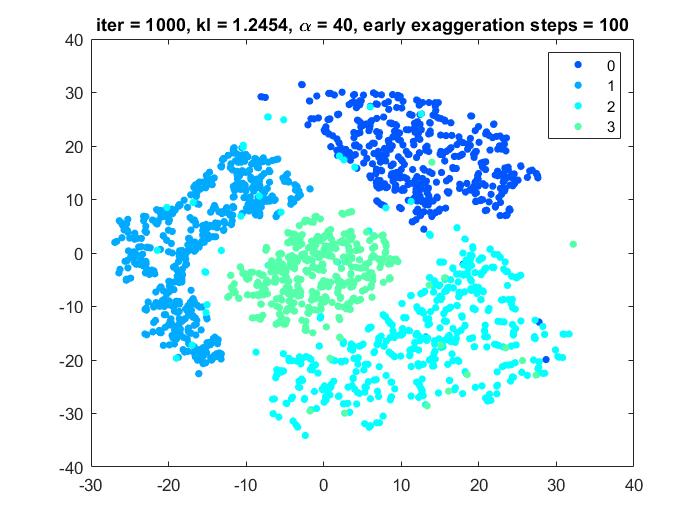}  
    \subcaption{ARS$_{2,2}$, $h = 1$, loss = 1.25}
    \end{subfigure}
     \begin{subfigure}{0.32\textwidth}
    \includegraphics[width=\textwidth,clip=true,trim=55 25 55 35]{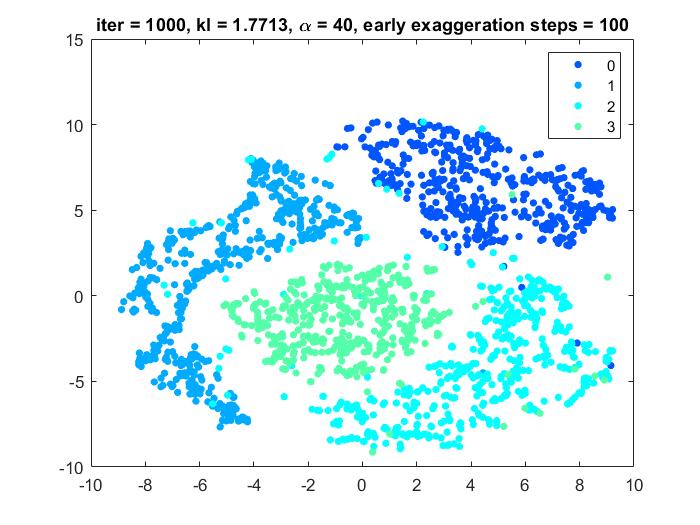}    
    \subcaption{t-SNE, $h = 70$, loss = 1.77}
    \end{subfigure}
     \begin{subfigure}{0.32\textwidth}
    \includegraphics[width=\textwidth,clip=true,trim=55 25 55 35]{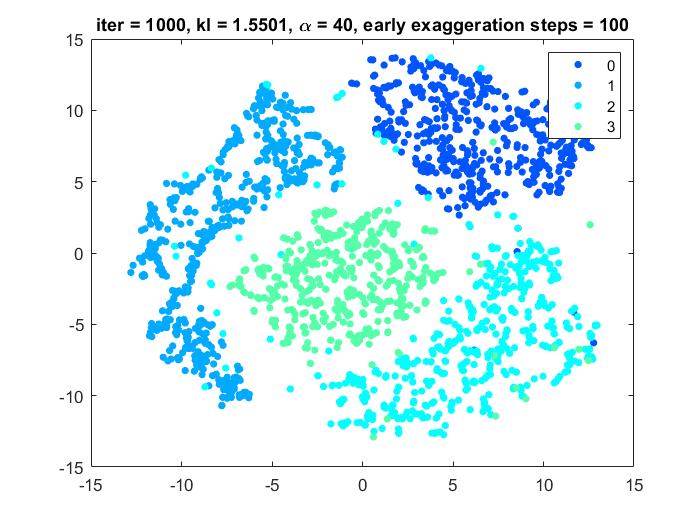}  
    \subcaption{t-SNE, $h = 150$, loss = 1.55}
    \end{subfigure}
    \caption{ARS$_{2,2}$ vs t-SNE applied to 2000 MNIST images of digits $0,1,2,3$, using $100$ steps of early exaggeration, $\a=40$. The top row corresponds to iteration $200$, while the bottom row is $1000$. For t-SNE, the time step $h$ is given for the gradient descent iterations, while $h=1$ is used during early exaggeration.}
    \label{fig: ARS vs GD N2000}
\end{figure}

The results of ARS and t-SNE under learning rate $h = 1$ after 200 and 1000 iterations are shown in Figure \ref{fig: ARS vs GD N1000}. We can see that ARS$_{2,2}$ achieves a good result, with sharp clusters, in only 200 iterations, while t-SNE under the same step size cannot converge sufficiently, even within 1000 iterations.  In fact, to achieve similar efficiency with ARS (based on the performance in the first 200 iterations), a much larger learning rate (around $h = 70$) is required for t-SNE during the gradient descent phase. Figure \ref{fig: ARS vs GD N2000} displays the same experiment when the data size is increased to 2000. We observe that our ARS method can still generate clear clusters within 200 iterations with $h = 1$, whereas the required learning rate for t-SNE is increased to around $h = 150$ in order to match the efficiency of ARS. Furthermore, in both experiments, the ARS algorithm yields the smallest value for the loss --- the Kullback-Leibler divergence --- even though the ARS algorithm is not designed to minimize this.

These results indicate that while the appropriate learning rate for t-SNE increases with the data size, our new ARS method can achieve good results with a constant $O(1)$ learning rate, due to our normalization of the forcing terms. The convenient choice of $h=1$ for the ARS learning rate will be further verified against larger data sets in Section \ref{sec: ARS-BH}. 

\begin{figure}[t!]
    \centering
    \begin{subfigure}{0.4\textwidth}
     \includegraphics[width=\textwidth,clip=true,trim=65 25 65 36]{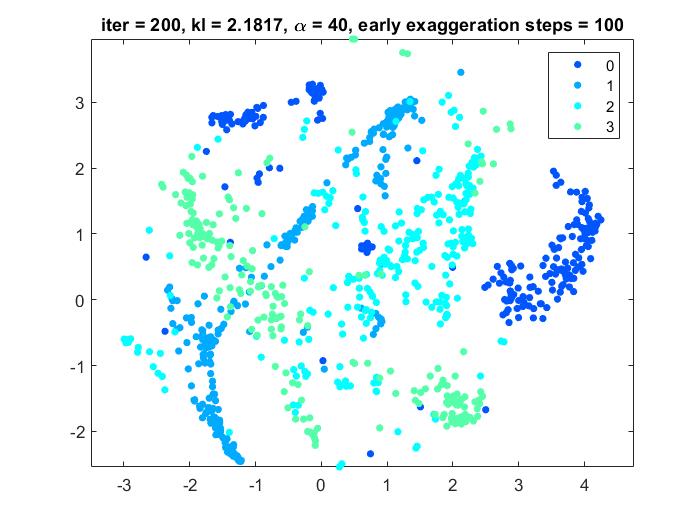}  
     \subcaption{iter = 200, loss = 2.1817}
    \end{subfigure}
    \quad
    \begin{subfigure}{0.4\textwidth}
    \includegraphics[width=\textwidth,clip=true,trim=65 25 65 36]{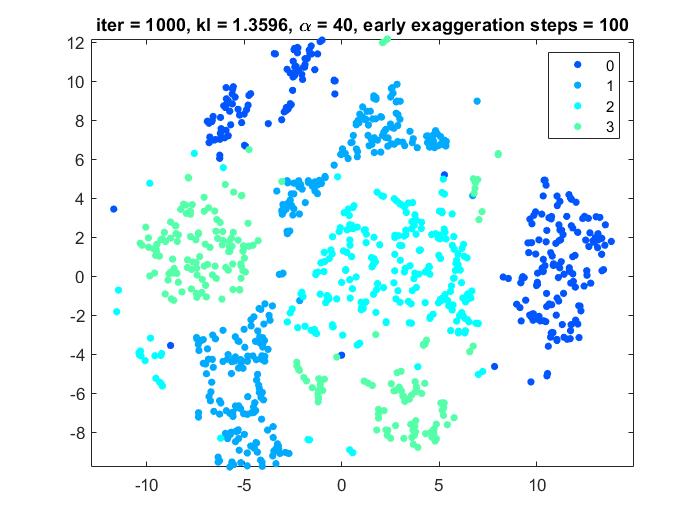}
     \subcaption{iter = 1000, loss = 1.3569}
    \end{subfigure}
    \caption{t-SNE with uniform learning rate $h = 70$ for both early exaggeration and gradient descent.}
    \label{fig: tSNE h = 70 for EE and GD}
\end{figure}

\begin{remark}
In the above computations, the learning rate for t-SNE is only increased during the gradient descent phase, whereas early exaggeration is still implemented with $h = 1$, to be consistent with the ARS test cases. Conventional implementations of t-SNE usually employ a consistent step size protocol for early exaggeration and gradient descent iterations. Nevertheless, we observe that applying a smaller learning rate for early exaggeration may improve the initial configuration. Figure \ref{fig: tSNE h = 70 for EE and GD} displays the result of t-SNE gradient descent under the same settings as those in Figures \ref{fig: ARS vs GD N1000} (e),(f), except that $h = 70$ is employed for \emph{both} early exaggeration and gradient descent. Here, an excessively large learning rate undermines the effectiveness of early exaggeration.
\end{remark}

\section{Tunable Interactions}\label{sec:tune}

The new attraction-repulsion swarming \eqref{eq: ARS} extends the framework of t-SNE by allowing user-tunable interaction protocols. We illustrate the benefits of such generalization through computational and analytical justifications.

\subsection{ARS$_{\th_1,\th_2}$} \label{sec:compare interactions}

 The ARS dynamics is dictated by parameters $\th_1$ and $\th_2$, as they control the localization of attraction and repulsive forcing. To demonstrate the influence of adjusting interaction kernels, we compare the performance of ARS$_{\th_1\th_2}$ under different settings against the same 1500 MNIST images with digits $0\sim 3$. All the test cases employ common initial data generated with uniform distribution $\by^0_i \sim \mathcal{U}([0,1]^2)$.

\begin{figure}[t!]
    \centering
    \begin{subfigure}{0.4\textwidth}
    \includegraphics[width=\textwidth,clip=true,trim=48 25 65 34]{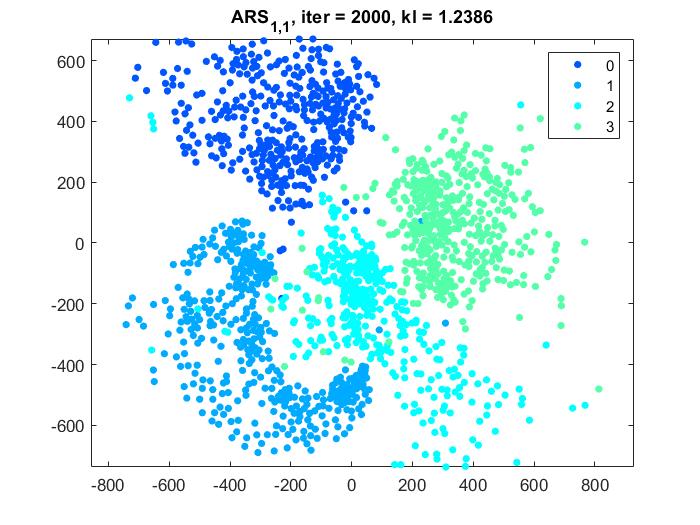} 
    \subcaption{ARS$_{1,1}$, loss = 1.2386}
    \end{subfigure}
    \quad
    \begin{subfigure}{0.4\textwidth}
    \includegraphics[width=\textwidth,clip=true,trim=48 25 65 34]{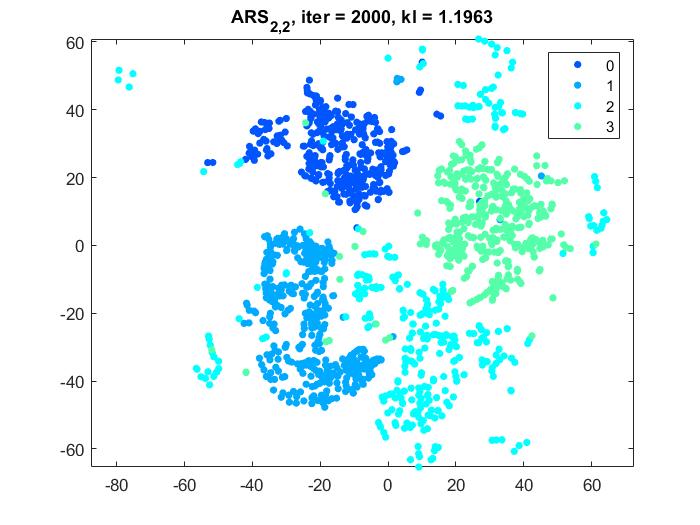} 
    \subcaption{ARS$_{2,2}$, loss = 1.1963}
    \end{subfigure}
    \\
    \begin{subfigure}{0.4\textwidth}
    \includegraphics[width=\textwidth,clip=true,trim=48 25 65 34]{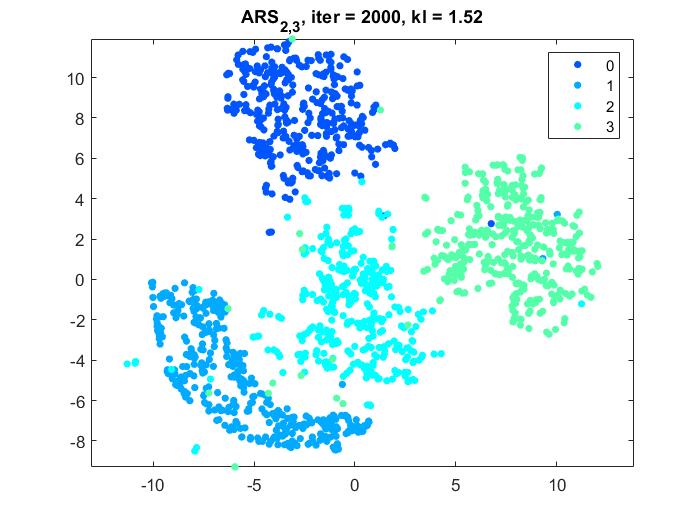} 
    \subcaption{ARS$_{2,3}$, loss = 1.5200}
    \end{subfigure}
    \quad
    \begin{subfigure}{0.4\textwidth}
    \includegraphics[width=\textwidth,clip=true,trim=48 25 65 34]{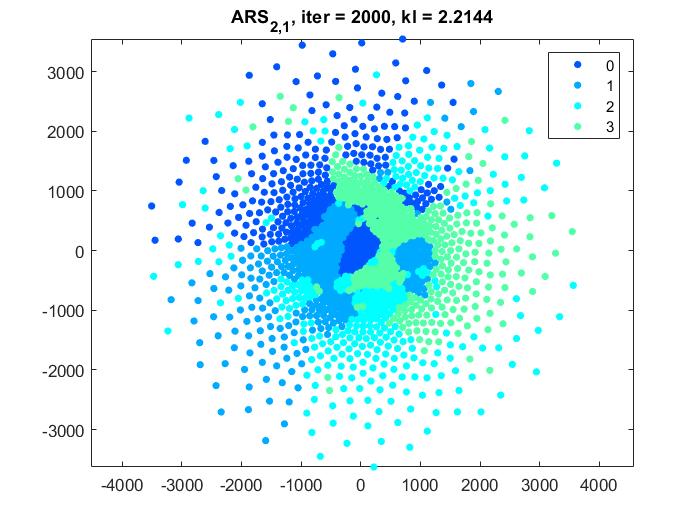} 
    \subcaption{ARS$_{2,1}$, loss = 2.2144}
    \end{subfigure}
    \caption{ARS$_{\th_1,\th_2}$ with different interaction kernels, 1000 MNIST data with digits 0,1,2,3, learning rate $h = 1$ and $2000$ iterations.}
    \label{fig:ARS th}
\end{figure}

Figure \ref{fig:ARS th} displays the results after 2000 iterations \emph{without} early exaggeration. It is seen that under $\th_1 = \th_2 = 2$ there are relatively many outliers with digit = 2 distributed away from the major cluster. This is due to the strong repulsion exerted by the other clusters blocking in between. However, with a greater value of $\th_2$, in which case the repulsive force is weakened at far-field, the quality of clustering is obviously improved with ARS$_{2,3}$. In fact, if the repulsive force is too strong at long distance in the sense $\th_2<\th_1$, the dynamics would fail to cluster the data, as shown by the result of ARS$_{2,1}$. Therefore, $\th_1\leq \th_2$ is required for practical implementation, namely, the repulsion interaction must be more localized than the attraction.

It is also worth noting that ARS$_{2,3}$ generates the most clear visualization among all the test cases although it does not admit the lowest loss. This implies that the Kullback-Leibler divergence may not be the best indicator to describe the quality of data visualization universally.

\subsection{Investigation of Localized Repulsive Forces}\label{sec:local repulsion}
The computations in Section \ref{sec:compare interactions} indicate the advantage of employing a localized repulsive force in terms of forming complete and well-separated clusters. In this section, we further study the effect of short-range repulsion.

\subsubsection*{Why favor localized repulsion?}

The benefits of using localized repulsive force can be interpreted from several aspects. From the computational point of view, localizing the repulsion dynamics would emphasize the influence of attraction at initial stage where the agents $\by_i$'s are loosely distributed. Such mechanism mimics the early exaggeration steps in the original t-SNE algorithm, which reinforces the communications between `similar' agents and accelerates the separation of distinct clusters afterwards. Meanwhile, by favoring attraction forces at long distance, the separated sub-clusters would have a greater chance of merging into one complete cluster, as observed from the comparison of ARS$_{2,2}$ and ARS$_{2,3}$ in Figure \ref{fig:ARS th}.

Applying a light-tail repulsion can also help prevent an excessively sparse distribution of agents $\by_i$'s. For convenience of notation, we denote the \emph{collective force} coefficient as
\[
\a_{ij}(t) := \tP_{ij}\psi_1(|\by_i(t)-\by_j(t)|)-\tQ_{ij}(t)\psi_2(|\by_i(t)-\by_j(t)|).
\]
Then the dynamics \eqref{eq:ARS dynamics} can be written as 
\begin{equation}
\frac{d}{dt} \by_i(t) = \sum^N_{j=1} \a_{ij}(t)(\by_j(t)-\by_i(t)).
\end{equation}

The influence of collective forcing on the crowd expansion is stated through the following proposition.

\begin{proposition} 
Let $\by_i(t)$, $i = 1,2,\cdots,N$ be the solution to the attraction-repulsion dynamics \eqref{eq:ARS dynamics} subject to initial data $\by_i(0) = \by^0_i$. Then the spatial diameter of the solution, 
\[
d_Y(t) := \max_{1\leq i,j \leq N}|\by_i(t)-\by_j(t)|, \quad t\geq 0,
\]
satisfies 
\begin{equation}\label{eq:diameter growth}
    -\max_{i\neq j}\a_{ij}(t)\cdot N \cdot d_Y(t) \leq \frac{d}{dt}d_Y(t) \leq -\min_{i\neq j}\a_{ij}(t)\cdot N\cdot d_Y(t),
\end{equation}
 
\end{proposition}

\noindent \emph{Proof.} We first prove the second inequality of the bound \eqref{eq:diameter growth}.  Arguing along the lines of \cite{HT2017,HT2020}, we fix an arbitrary unit vector $\bw\in\R^2$, the projection of \eqref{eq:ARS dynamics} onto the space spanned by $\bw$ yields
\begin{equation*}
    \frac{d}{dt} \lan\by_i(t),\bw\ran = \sum^N_{j = 1} \a_{ij}(t)(\lan\by_j,\bw\ran-\lan\by_i,\bw\ran).
\end{equation*}
Assume that $\lan\by_i(t),\bw\ran$ reaches a maximum at $i = i^+(t)$ and a minimum at $i = i^-(t)$. Denote
\[
y^+(t) := \max_{1\leq i\leq N}\lan\by_i(t),\bw\ran = \lan \by_{i^+(t)}(t),\bw\ran.
\]
Computation of the time evolution of $y^+(t)$ yields
\begin{equation}\label{eq:y+ bound}
\begin{split}
    \frac{d}{dt}y^+(t) &= \sum^N_{j=1}\a_{i^+ j}(t)(\lan\by_j(t),\bw\ran-y^+(t))\\
    &\leq -\min_{i \neq j}\a_{ij}(t)\cdot\sum^N_{j=1} (y^+(t)-\lan\by_j(t),\bw\ran)
\end{split}
\end{equation}
Similarly, the time evolution of $\displaystyle y^-(t):= \min_{1\leq i\leq N}\lan\by_i(t),\bw\ran$ can be estimated with
\begin{equation}\label{eq:y- bound}
    \frac{d}{dt} y^-(t) \geq \min_{i,j} \a_{ij}(t) \cdot\sum^N_{j=1} (\lan\by_j(t),\bw\ran-y^-(t)).
\end{equation}
The difference of \eqref{eq:y+ bound} and \eqref{eq:y- bound} yields
\begin{equation}
    \frac{d}{dt} (y^+(t)-y^-(t)) \leq -\min_{i \neq j}\a_{ij}(t)\cdot N \cdot(y^+(t)-y^-(t)).
\end{equation}
Since the above inequality holds for an arbitrary unit vector $\bw$, it follows that the diameter of the solution, $\displaystyle d_Y(t) = \sup_{|\bw|=1}\max_{1\leq i,j\leq N}\lan \by_i(t)-\by_j(t),\bw\ran$, satisfies
\[
\frac{d}{dt}d_Y(t) \leq -\min_{i \neq j}\a_{ij}(t) \cdot N \cdot d_Y(t).
\]
This concludes the desired estimate. The first inequality of \eqref{eq:diameter growth} can be obtained along similar argument. $\hfill \square$

The bound \eqref{eq:diameter growth} indicates that the expansion/contraction of the crowd is the result of the competition between attraction and repulsion forces. The case of $\a_{ij}(t)>0$ corresponds to the situation where the interaction between $\by_i$ and $\by_j$ is dominated by attraction, whereas $\a_{ij}(t)<0$ indicates the dominance of repulsion force.

Assuming the attraction kernel $\psi_1$ admits a heavier tail than the repulsion kernel $\psi_2$ in the sense that $\lim_{r\rightarrow \infty} \frac{\psi_1(r)}{\psi_2(r)} = \infty$ (e.g., $\psi_i(r) = (1+r^{\th_i})^{-1}$, $i = 1,2$, with $\th_1<\th_2$), then there exists $R_+>0$ such that 
\[
\frac{\psi_1(r)}{\psi_2(r)} > \frac{1}{\min_{i\neq j}\tP_{ij}}>0, \quad \forall r>R_+.
\]
Thus, employing thin-tail (localized) repulsion helps avoid \emph{over-expansion}, $|\by_i-\by_j|>R_+$ for $\forall i,j$ --- indeed, once over-expansion occurs, the attraction effect becomes uniformly dominating,
\[
\a_{ij}(t) \geq \tP_{ij}\psi_1(|\by_i-\by_j|)-\psi_2(|\by_i-\by_j|) > \psi_2(|\by_i-\by_j|)\left(\frac{\tP_{ij}}{\min_{k \neq l}\tP_{kl}}-1\right)>0, \quad \forall  i \neq j.
\]
The first line was obtained due to the fact that the repulsion amplitude remains uniformly bounded, $\tQ_{ij}(t) \leq 1$ for $\forall i,j$ and $t\geq 0$. Consequently, the bound \eqref{eq:diameter growth} implies global contraction of the crowd, $\frac{d}{dt}d_Y(t) < 0$. 

The situation of \emph{over-contraction} $d_Y(t) \ll 1$, however, is more complicated. Noting that $\lim_{d_Y\rightarrow 0}\tQ_{ij} = \frac{1}{N-1}$, $i\neq j$, we have
\[
\lim_{d_Y\rightarrow 0}\a_{ij} = \tP_{ij}\psi_1(0+)-\frac{1}{N-1}\psi_2(0+), \quad i\neq j.
\]
If the repulsion kernel has a heavier head than the attraction kernel in the sense that $\lim_{r\rightarrow 0} \frac{\psi_2(r)}{\psi_1(r)} = \infty$ (e.g., $\psi_1 = (1+r^{\th_1})^{-1}$, $\th_1>0$ and $\psi_2(r) = r^{-s}$, $s>0$), then there exists $R_->0$ such that 
\[
\frac{\psi_2(r)}{\psi_1(r)}>(N-1)\cdot\max_{i\neq j}\tP_{ij}, \quad \forall r<R_-.
\]
Once the distribution of $\by_i$'s becomes overly crowded $d_Y(t) < R_-$, the heavy-head kernel $\psi_2$ leads to the uniform dominance of repulsion, $\a_{ij}(t)<0$ for $\forall i\neq j$. Hence, the bound \eqref{eq:diameter growth} implies the global expansion, $\frac{d}{dt} d_Y(t)>0$. This prevents the occurrence of the trivial solution, $\by_i = \by_j$, $\forall i,j = 1,\cdots,N$, as long as the initial data is non-uniform.

However, according to our experiments, an excessively strong repulsive force, even within a short range only, can result in the failure of cluster formation. Therefore, for actual computations, we recommend to use repulsion kernels with moderate head, $\displaystyle \frac{\psi_2(r)}{\psi_1(r)} \approx 1$, $r\rightarrow 0$. In this case, the short-range interaction between agents $\by_i$ and $\by_j$ becomes more subtle --- if $\by_i$ and $\by_j$ belong to the same cluster,  we may expect $\tP_{ij} \approx 1$. Then the attraction effect  may dominate their interaction at all distances, which is acceptable for data clustering. On the other hand, if the two agents have sufficiently distinct features $\tP_{ij}\ll 1$, then the repulsive force can be expected to dominate at a short distance, even if its head is not so heavy. Such a mechanism would prevent the degeneration of the global configuration into a single point. The analytical proof will be left for future study. 

\subsubsection*{How much should repulsion be localized?}

Given the observed advantage of thin-tail repulsion, it is of interest to explore to what extent the repulsive force should be localized. To this end, we compare the performances of different repulsion kernels against the same 2000 MNIST images with digits $0\sim 6$. All the test cases employ the same initial data generated with the uniform distribution $\by^0_i \sim \mathcal{U}([0,1]^2)$. No early exaggeration steps are implemented in computations. The attraction kernel is fixed to $\psi_1(r) = (1+r^2)^{-1}$ and perplexity is set to 30.

\begin{figure}[t!]
    \centering
    \begin{subfigure}{0.32\textwidth}
    \includegraphics[width=\textwidth,clip=true,trim=65 25 55 34]{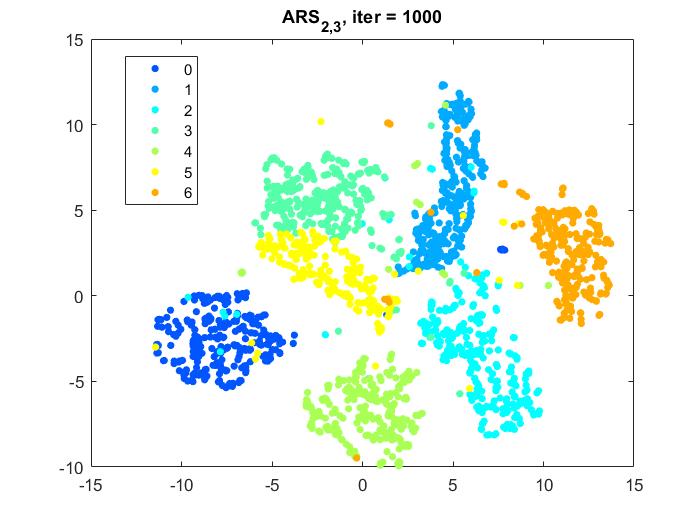}    
    \subcaption{ARS$_{2,3}$}
    \end{subfigure}
    \begin{subfigure}{0.32\textwidth}
    \includegraphics[width=\textwidth,clip=true,trim=65 25 55 34]{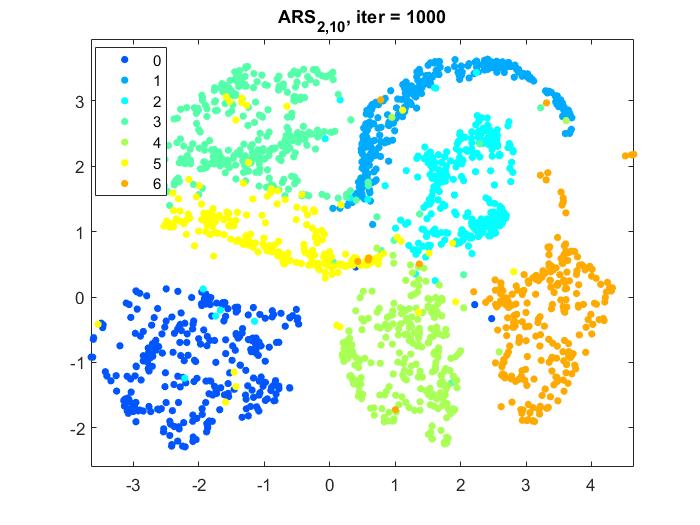}    
    \subcaption{ARS$_{2,10}$}
    \end{subfigure}
     \begin{subfigure}{0.32\textwidth}
    \includegraphics[width=\textwidth,clip=true,trim=65 25 55 34]{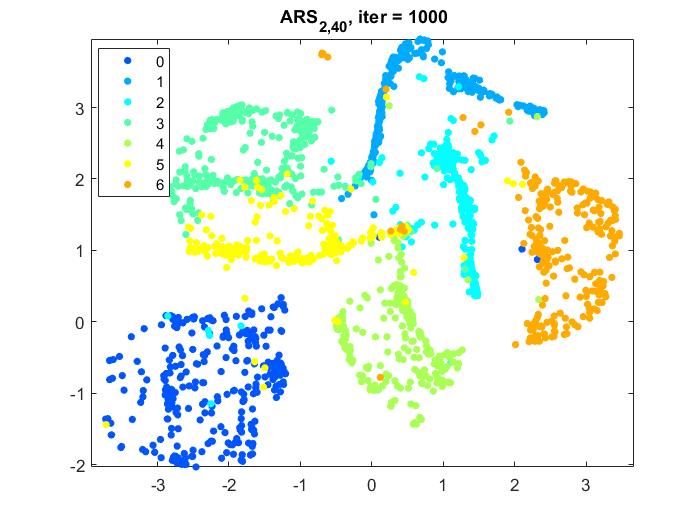}    
    \subcaption{ARS$_{2,40}$}
    \end{subfigure}
    \caption{ARS with polynomial decaying repulsion, 2000 MNIST data with digits 0$\sim$6, 1000 steps of iterations, learning rate of $h=1$.}
    \label{fig:ARS th N2000 labels 0 to 6}
\end{figure}

Figure \ref{fig:ARS th N2000 labels 0 to 6} displays the results of polynomial decaying repulsion kernels, $\psi_2(r) = (1+r^{\th_2})^{-1}$, under different values of $\th_2$ after 1000 iterations. We observe that for greater values of $\th_2$, the configuration becomes more `crowded', i.e., the distances between different clusters are smaller, due to the reduced repulsive force at long distance. Nevertheless, all the test cases seem to cluster the data points reasonably well.

\begin{figure}[t!]
    \centering
    \begin{subfigure}{0.32\textwidth}
    \includegraphics[width=\textwidth,clip=true,trim=34 25 55 38]{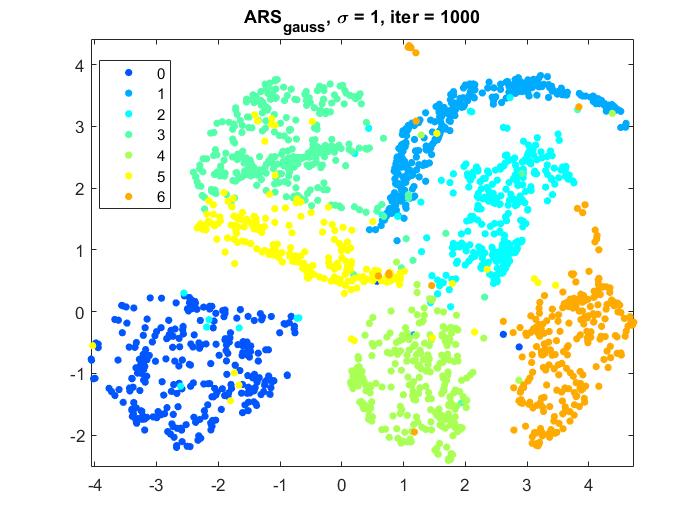}
    \subcaption{ARS$_{Gauss}$, $\sig = 1$}
    \end{subfigure}
    \begin{subfigure}{0.32\textwidth}
    \includegraphics[width=\textwidth,clip=true,trim=34 25 55 38]{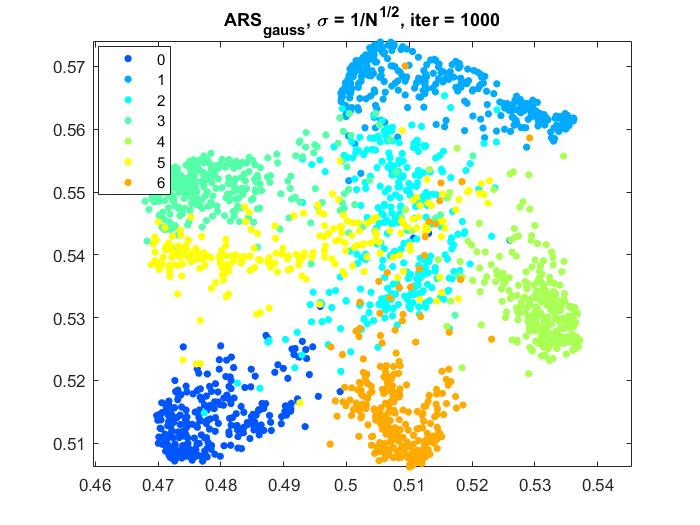}
    \subcaption{ARS$_{Gauss}$, $\sig = 1/\sqrt{N}$}
    \end{subfigure}
    \begin{subfigure}{0.32\textwidth}
    \includegraphics[width=\textwidth,clip=true,trim=34 25 55 38]{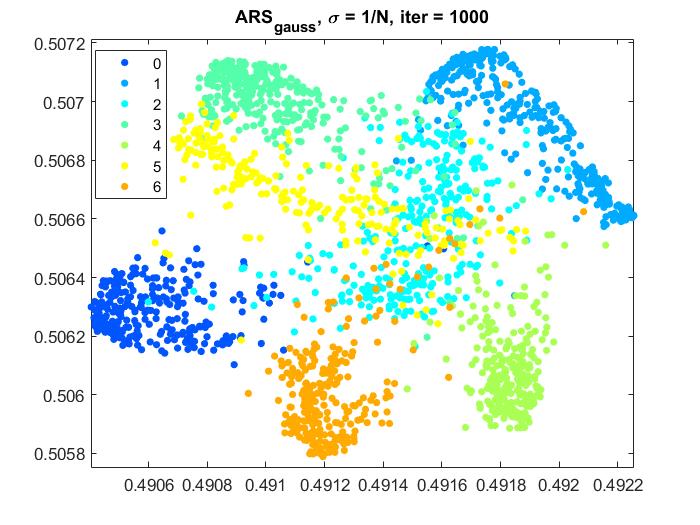}
    \subcaption{ARS$_{Gauss}$, $\sig = 1/N$}
    \end{subfigure}
    \caption{ARS with Gaussian repulsion, 2000 MNIST data with labels 0$\sim$6, 1000 steps of iterations, learning rate of $h=1$.}
    \label{fig:ARS gauss N2000 labels 0 to 6}
\end{figure}

We also investigate the use of the Gaussian repulsion kernel in ARS (ARS$_{Gauss}$), which admits even thinner tails, $\psi_2(r) = \exp(-r^2/\sig^2)$, $\sig>0$. With a smaller value of $\sig$, the interaction is more concentrated near the center $r = 0$. The numerical results of ARS$_{Gauss}$ are shown in Figure \ref{fig:ARS gauss N2000 labels 0 to 6}. It is seen that for excessively small values of $\sig$, the clusters are overly crowded. Indeed, since the repulsion force is globally too weak, different clusters are much less distinguishable and greatly overlap with each other, which deteriorates the visualization.

Overall, despite the benefits of using shorter-range repulsion, we should still avoid over-localization in order to keep appropriate distance between distinct clusters. Under such consideration, the polynomial decay repulsion kernels have better performance than Gaussian repulsion kernels. In particular, ARS$_{2,3}$ seems to give the best visualization universally.

\section{Mean-Field Limit of ARS}\label{sec:meanfield}

The analytical discussions in Section \ref{sec:local repulsion} primarily demonstrate the effect of force balancing on the particle swarming process. Nevertheless, the inter-connection between the structure of high dimensional input data and the clustering behavior of the low-dimensional embedding still lacks theoretical guarantees. To reinforce the interpretability of ARS results, especially when a large data size is involved, studying the mean-field limit may provide a tractable approach. We give an outline of a possible approach here, leaving the analysis to future work. For simplicity, we assume a constant bandwidth, $\sig_i \equiv \sig$, is employed for all data points.

To obtain the mean-field description of the ARS particle system \eqref{eq: ARS}, we define the empirical distributions of the input data and output projections,
\[
\mu(\bx) = \frac{1}{N}\sum^N_{i=1}\delta_{\bx_i}(\bx), \quad \bx \in \R^D, \ \ \text{and} \ \ \rho_t(\by) = \frac{1}{N}\sum^N_{i=1}\delta_{\by_i(t)}(\by), \quad \by\in\R^d. 
\]
The measures $\mu$ and $\rho_t$ are connected through the mapping, $T^N_t: \R^D\mapsto\R^d$, with
\[
T^N_t(\bx_i) = \by_i(t), \quad 1\leq i \leq N.
\]
The time evolution of $T^N_t$ encodes the clustering process of the low-dimensional embedding $\{\by_i\}$. Let us assume that the discrete projection mapping admits a (formal) large-crowd limit,
\[
 T_t = \lim_{t\rightarrow\infty} T^N_t.
\]
Then the evolution of the mean-field measure, $\rho_t = T_t(\mu)$, is governed by the nonlinear transport equations
\begin{equation}\label{eq:ARS mean-field}
\left.\begin{array}{l}
 \partial_t \rho_t + \nabla\cdot(\rho_t \bu) = 0 \\
\displaystyle \bu(t,\by) = \int\int F(\by,\by',t)(\by'-\by)\rho_t(\by')d\by'
\end{array}
\right\}.
\end{equation}
The kernel function, $F(\by,\by',t)$, reflects the balance between attraction and repulsive forces,
\[
F(\by,\by',t) = \tP(\by,\by',t)\psi_1(|\by-\by'|)-\tQ(\by,\by',t)\psi_2(|\by-\by'|),
\]
with  $S_t(\by) = \{\bx\in \supp\{\mu\}: T_t (\bx) = \by\}$,
\[
\begin{split}
&\tP(\by,\by',t) = \frac{\iint_{S_t(\by)\times S_t(\by')} e^{-\frac{|\bx-\bx'|}{2\sig^2}}\mu(\bx)\mu(\bx')d\bx d\bx'}{\int_{\R^d}[\iint_{S_t(\by)\times S_t(\bz)} e^{-\frac{|\bx-\bx'|}{2\sig^2}}\mu(\bx)\mu(\bx')d\bx d\bx']\rho_t(\bz)d\bz}, \ \  \text{and}\\
& \tQ(\by,\by',t) = \frac{1}{\int_{R^d}\frac{1}{1+|\by-\bz|^2}\rho_t(\bz)d\bz}\cdot \frac{1}{1+|\by-\by'|^2}.
\end{split}
\]
For the derivation of the mean-field equations \eqref{eq:ARS mean-field} see Appendix \ref{appendix: mean-field}.

Although t-SNE has shown promising computational outcomes in many existing studies, the rigorous proof of cluster formation still remains a big challenge due to the complicated particle interactions within $N-$body system. Our new ARS model, due to the force normalization, admits a relatively convenient mean-field limit by averaging over a large crowd of agents.\footnote{Notice that the forcing term (gradient) in the original t-SNE algorithm scales with $O(1/N)$, which converges to a zero limit as $N\rightarrow\infty$. Hence the mean-field limit for the \emph{particle distribution} is not available. Nevertheless, a mean-field limit may be derived for the \emph{energy functional} \eqref{eq:KL divergence} under certain graph structures. For details, we refer to \cite{steinerberger2022t}.} With appropriate assumptions on the input data distribution $\mu$, which subsequently affects the temporal evolution of similarity $\tP(\by,\by',t)$, it may be possible to prove the time-asymptotic convergence of $\rho_t$ towards a limiting distribution, $\displaystyle \rho_\infty = \lim_{t\rightarrow\infty}\rho_t$.  We will leave this investigation for future study.

\section{Tree-Based Variant: Barnes-Hut ARS}\label{sec: ARS-BH}

The primary versions of t-SNE and ARS algorithms involve computing the interactions between $N(N-1)$ pairs of agents, as well as storing the $N\times N$ input similarities $P$ and projection similarities $Q$. The quadratic growth of computational complexity with respect to the data size restricts the use of t-SNE in large data visualization. Inspired by the Barnes-Hut algorithm \cite{barnes1986hierarchical} and the dual-tree algorithms \cite{gray2000n}, van der Maaten \cite{van2014accelerating} proposed a tree-based variant of t-SNE called Barnes-Hut t-SNE. The key components include {\bf 1.} employing a sparse approximation for the input similarity matrix $P$, and {\bf 2.} approximating the agent-agent repulsion dynamics with agent-cell or cell-cell interactions, which can be computed efficiently using tree data structures. In this way, the computational complexity of computing the t-SNE gradient can be reduced from $O(N^2)$ to $O(N\log N)$. We can follow a nearly identical development to approximate the ARS forcing terms in $O(N\log N)$ computational time, which we call Barnes-Hut ARS, or ARS-BH. We refer to  \cite{van2014accelerating} for more details on the Barnes-Hut approximation, and have made our code for ARS-BH available on GitHub,\footnote{See \url{https://github.com/jwcalder/AttractionRepulsionSwarming-tSNE}.} as well as part of the GraphLearning Python package \cite{calder2022graphlearningSOFTWARE}.

\begin{figure}[t!]
\centering
\begin{subfigure}{0.32\textwidth}
\includegraphics[width=\textwidth]{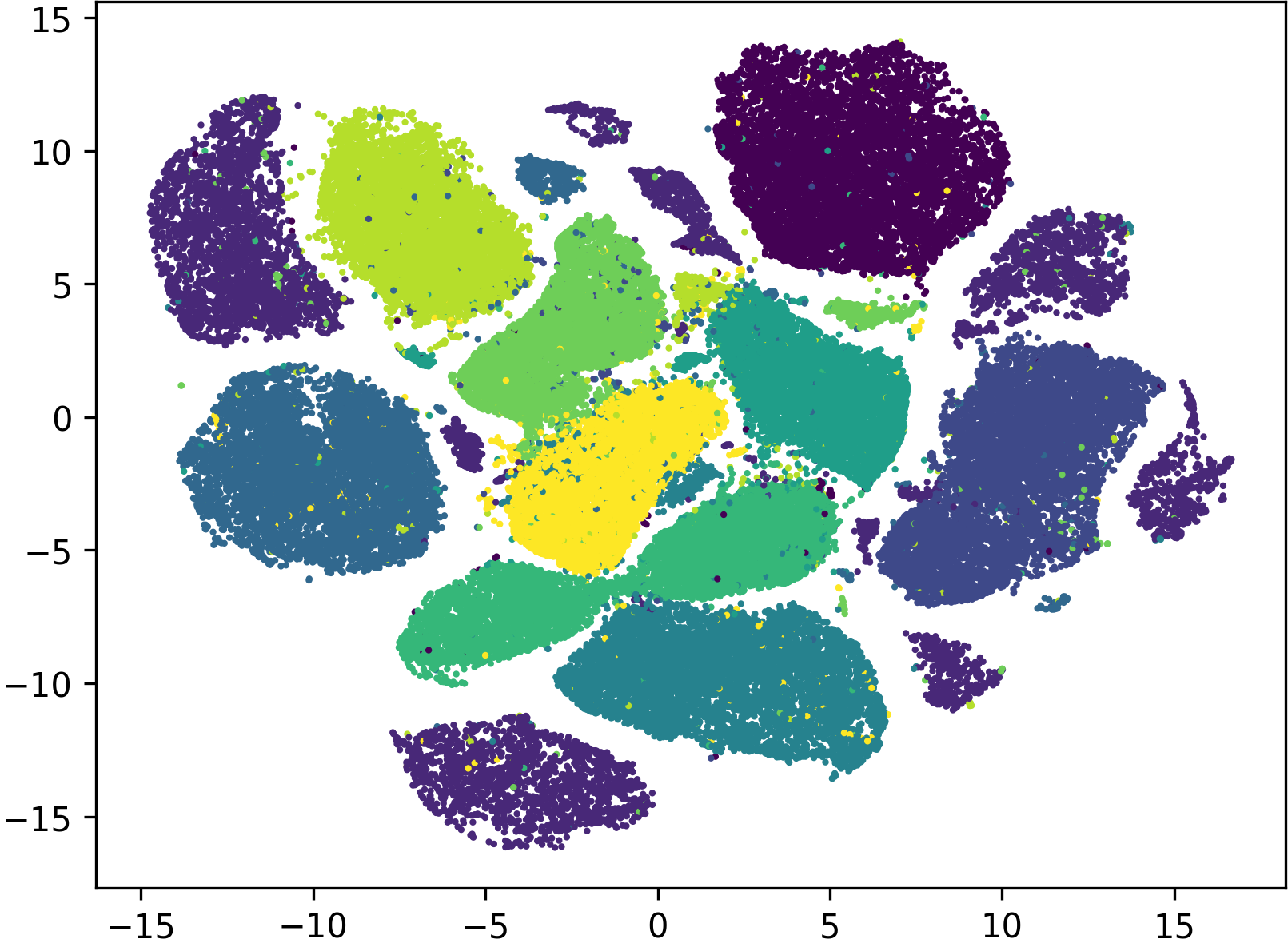}    
\subcaption{$\alpha=1, T=500$}
\end{subfigure}
\begin{subfigure}{0.32\textwidth}
\includegraphics[width=\textwidth]{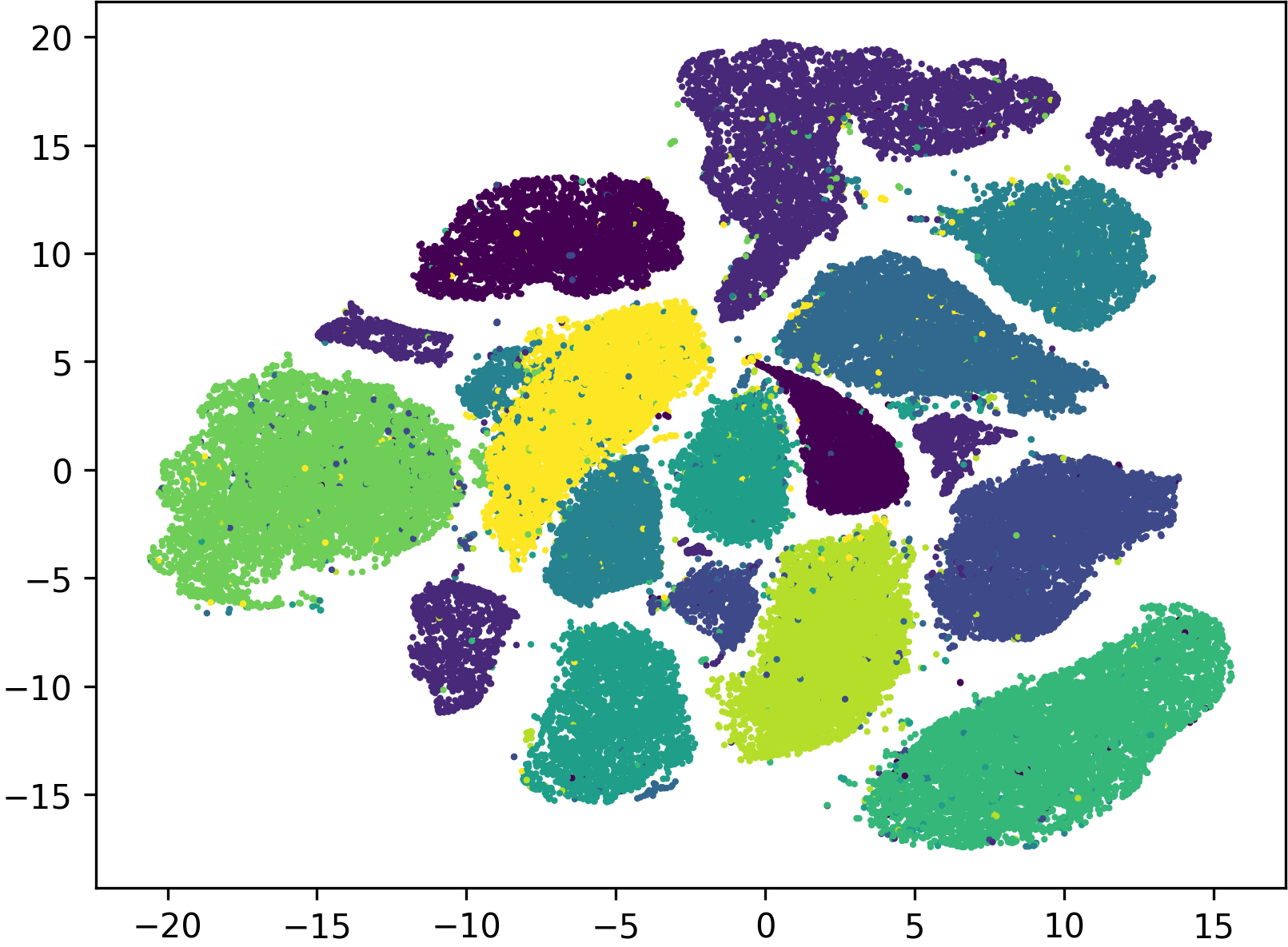}    
\subcaption{$\alpha=1, T=1000$}
\end{subfigure}
\begin{subfigure}{0.32\textwidth}
\includegraphics[width=\textwidth]{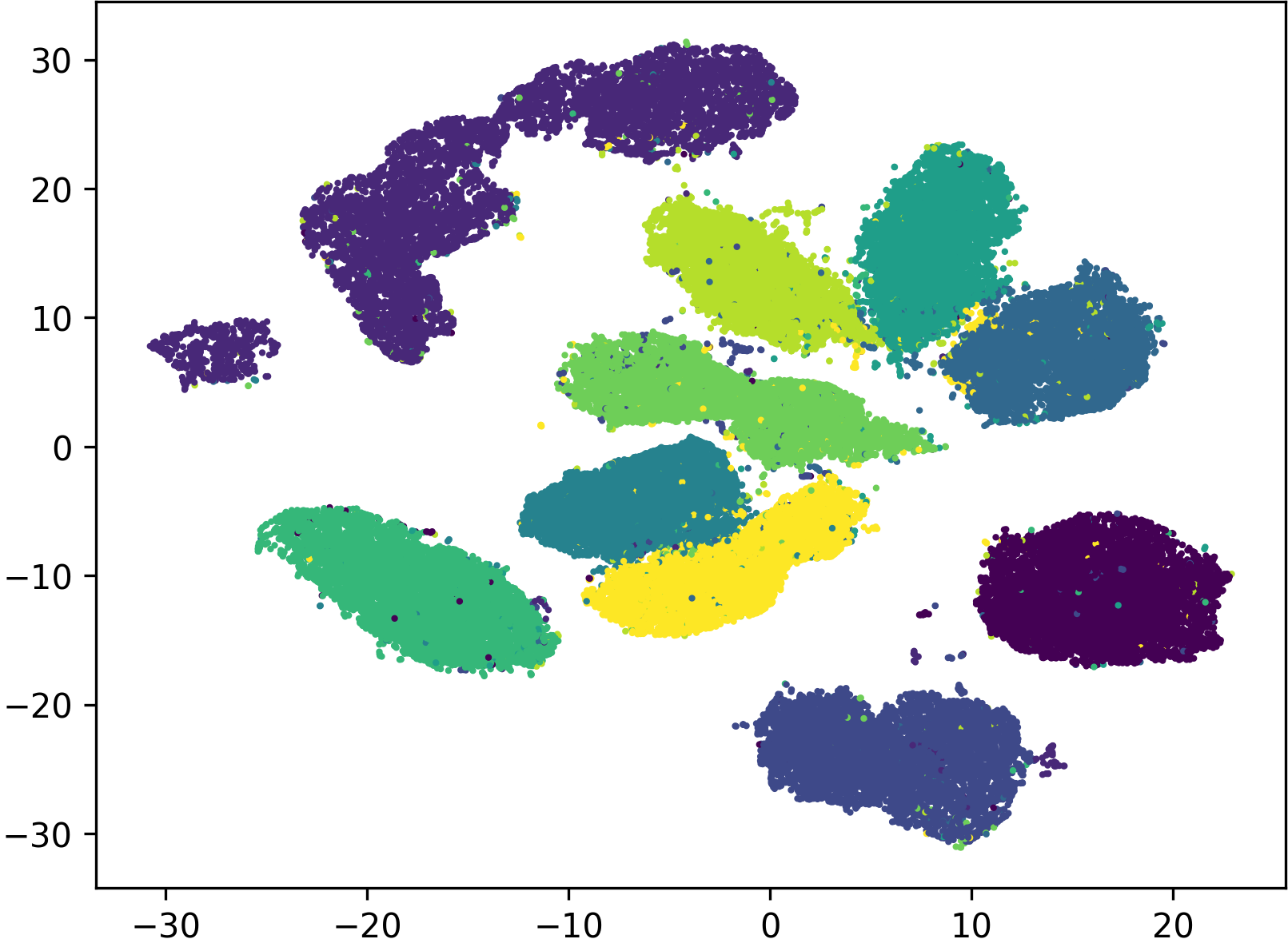}    
\subcaption{$\alpha=1, T=5000$}
\end{subfigure}\\
\begin{subfigure}{0.32\textwidth}
\includegraphics[width=\textwidth]{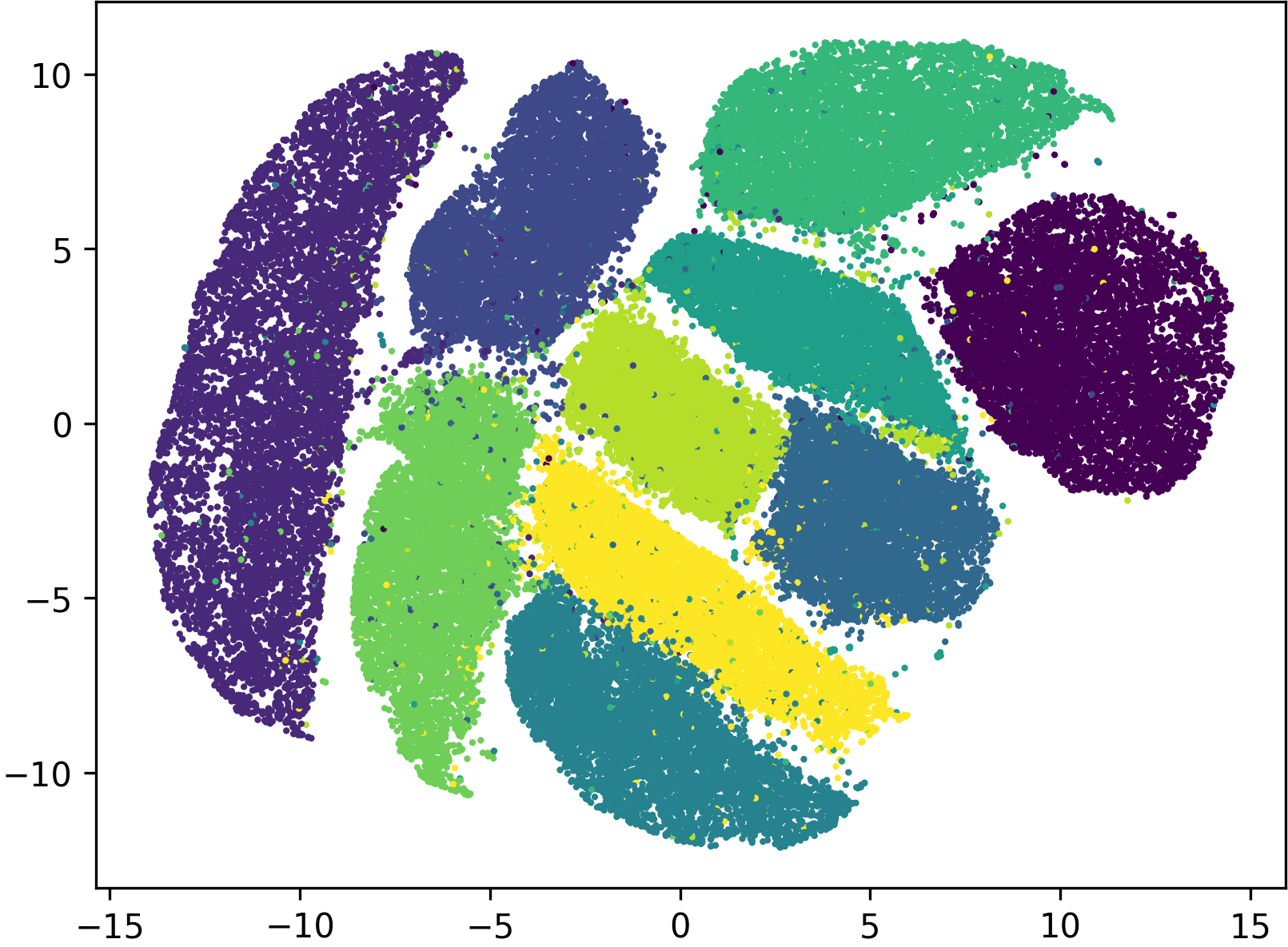}    
\subcaption{$\alpha=10, T=500$}
\end{subfigure}
\begin{subfigure}{0.32\textwidth}
\includegraphics[width=\textwidth]{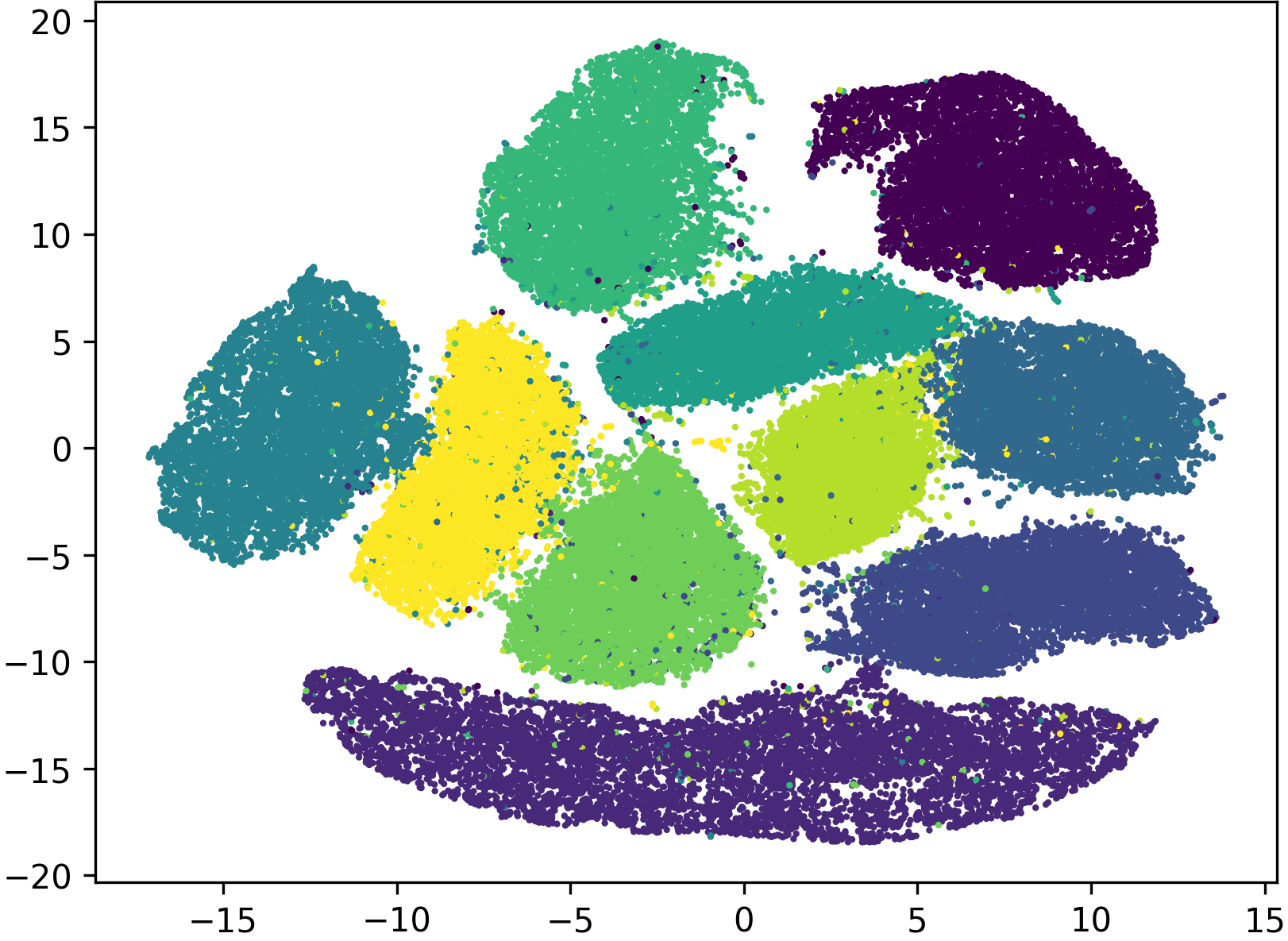}    
\subcaption{$\alpha=10, T=1000$}
\end{subfigure}
\begin{subfigure}{0.32\textwidth}
\includegraphics[width=\textwidth]{mnist_raw_2.0_3.0_10.0_5000.png}    
\subcaption{$\alpha=10, T=5000$}
\end{subfigure}
\caption{ARS-BH on MNIST with $\theta_1=2$, $\theta_2=3$ and $h=1$.}
\label{fig:bh_tsne_mnist}
\end{figure}

\begin{figure}[t!]
\centering
\begin{subfigure}{0.32\textwidth}
\includegraphics[width=\textwidth]{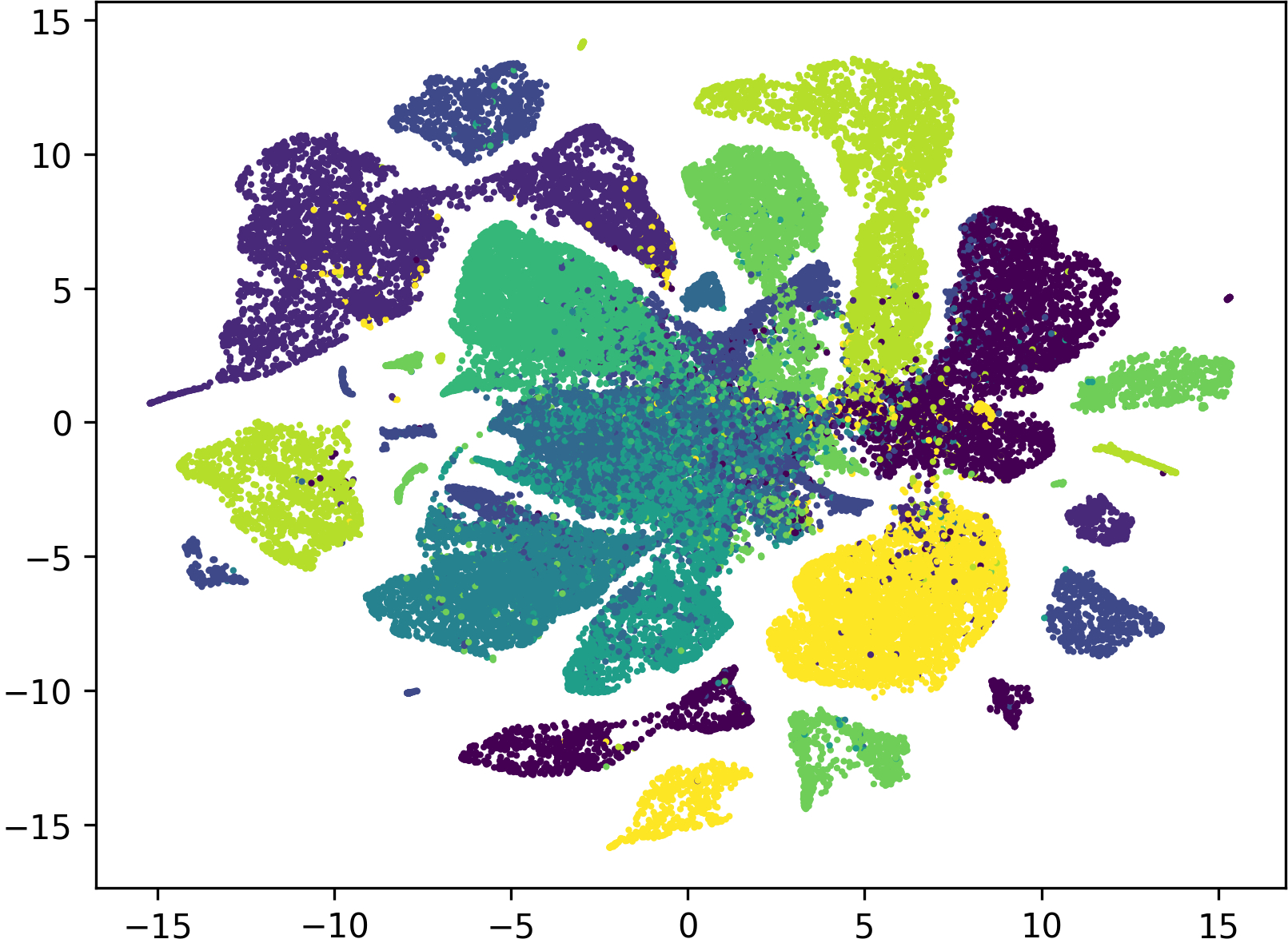}    
\subcaption{$\alpha=1, T=500$}
\end{subfigure}
\begin{subfigure}{0.32\textwidth}
\includegraphics[width=\textwidth]{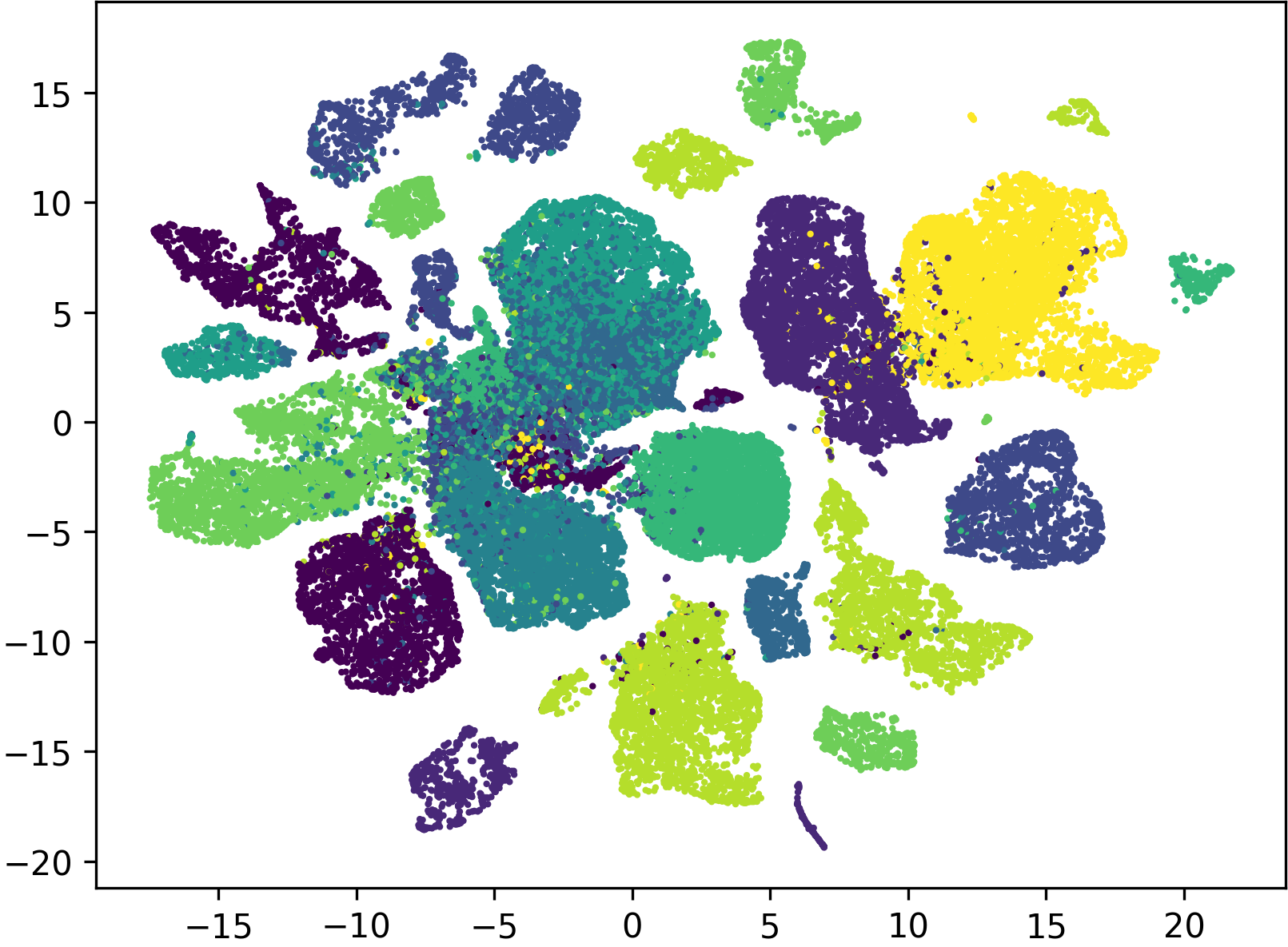}    
\subcaption{$\alpha=1, T=1000$}
\end{subfigure}
\begin{subfigure}{0.32\textwidth}
\includegraphics[width=\textwidth]{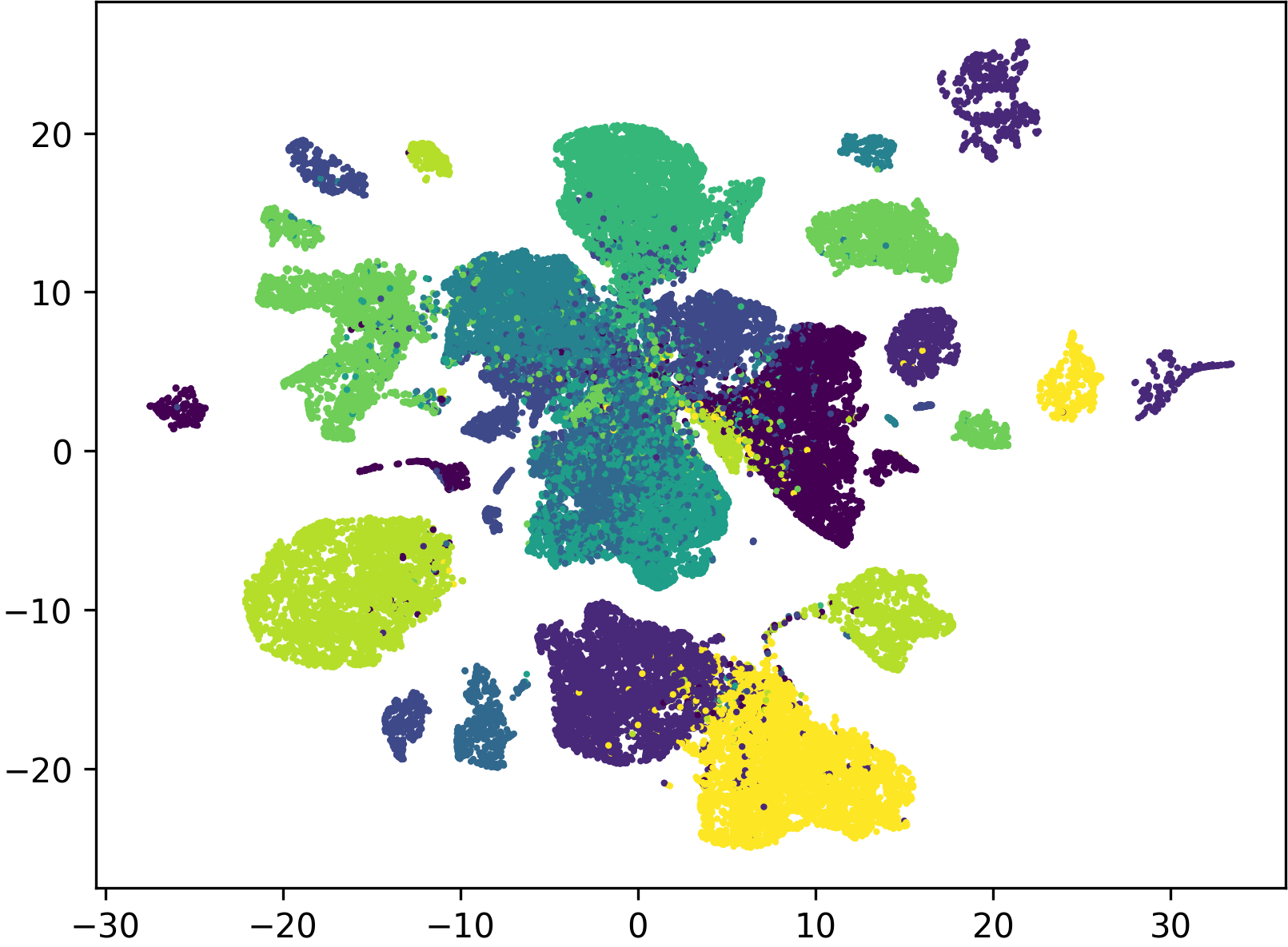}    
\subcaption{$\alpha=1, T=5000$}
\end{subfigure}\\
\begin{subfigure}{0.32\textwidth}
\includegraphics[width=\textwidth]{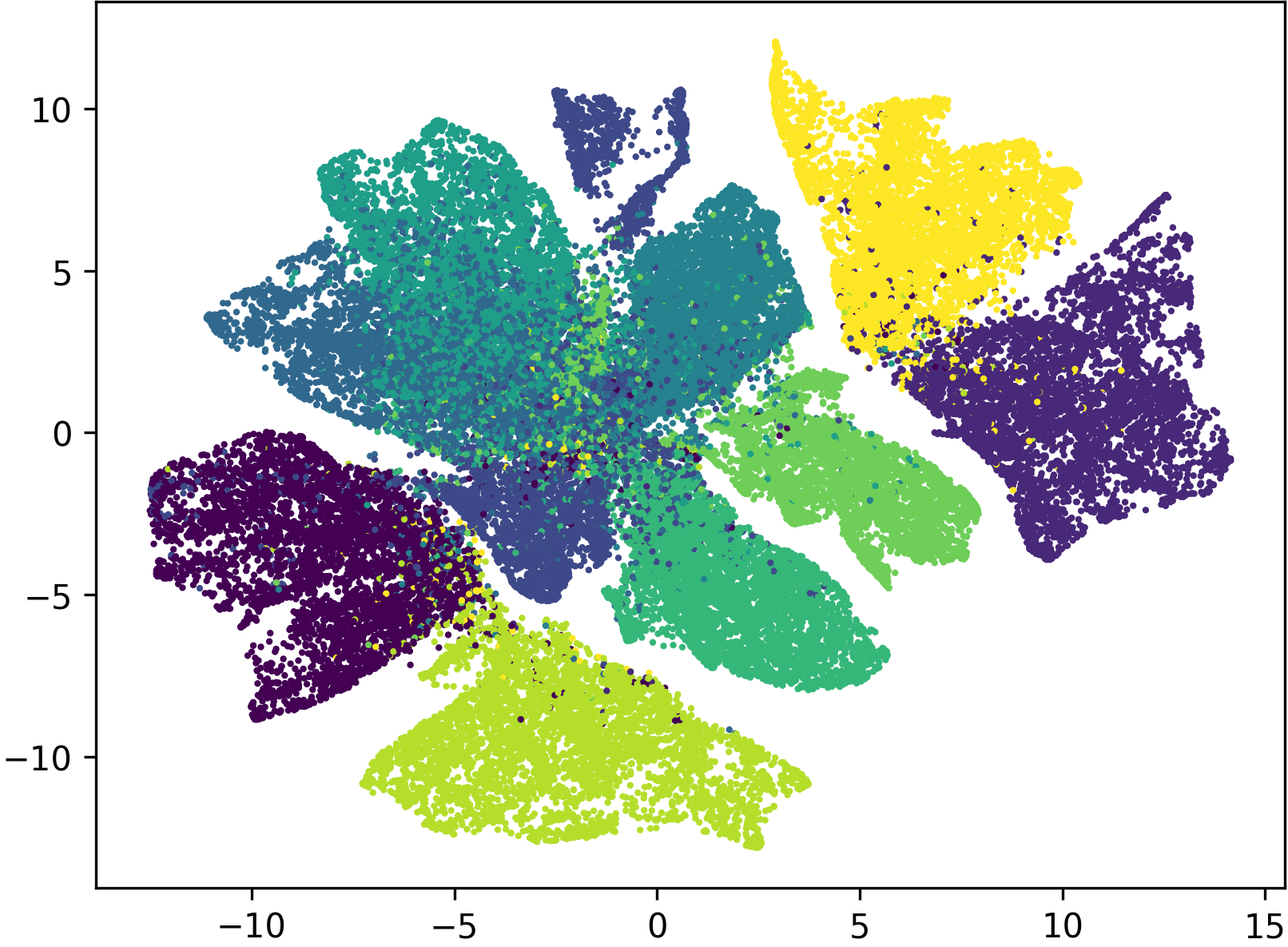}    
\subcaption{$\alpha=10, T=500$}
\end{subfigure}
\begin{subfigure}{0.32\textwidth}
\includegraphics[width=\textwidth]{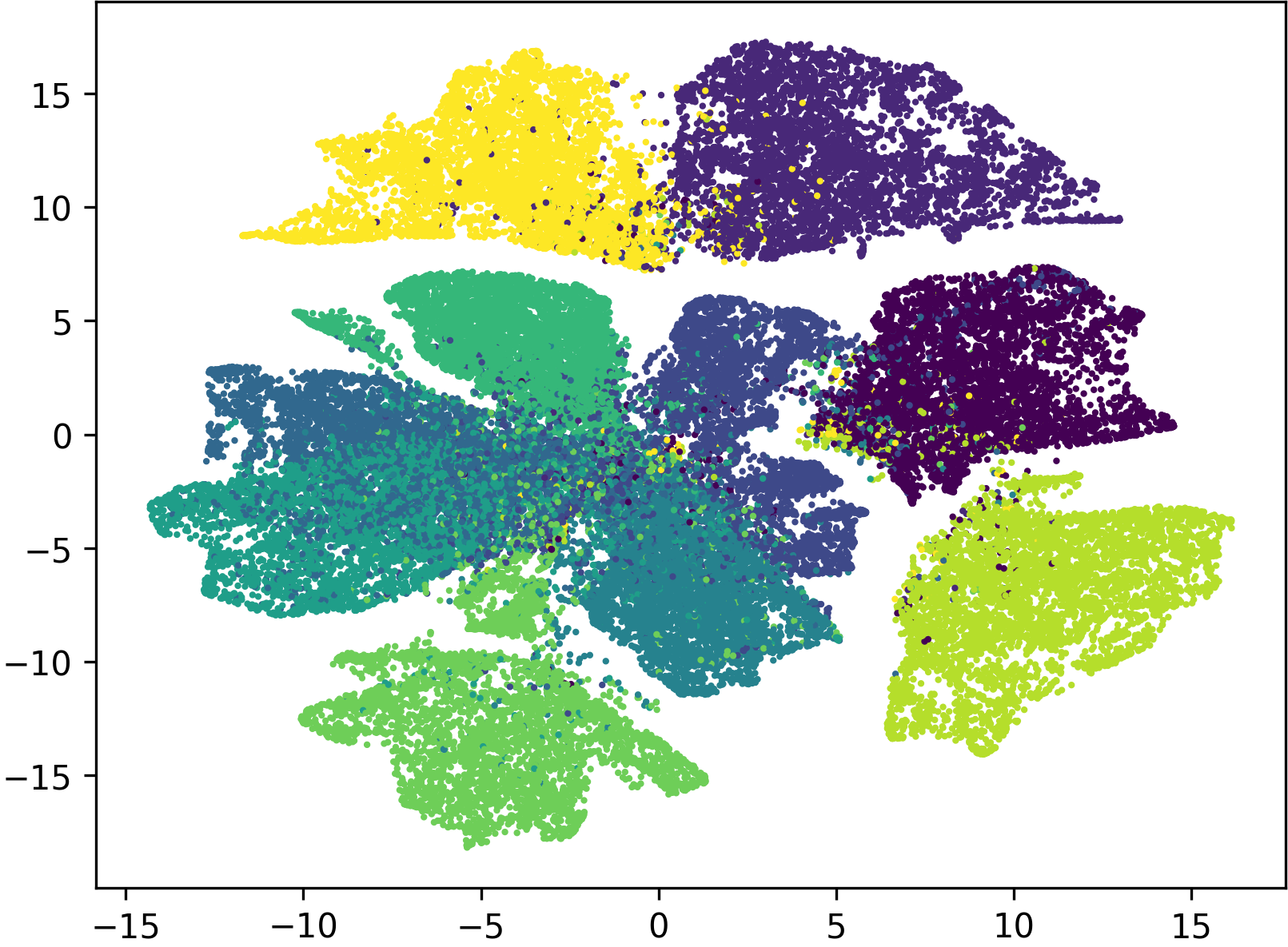}    
\subcaption{$\alpha=10, T=1000$}
\end{subfigure}
\begin{subfigure}{0.32\textwidth}
\includegraphics[width=\textwidth]{cifar10_simclr_2.0_3.0_10.0_5000.png}    
\subcaption{$\alpha=10, T=5000$}
\end{subfigure}
\caption{ARS-BH on Cifar-10 with $\theta_1=2$, $\theta_2=3$ and $h=1$.}
\label{fig:bh_tsne_cifar}
\end{figure}

Using the ARS-BH method, we can scale up the computations with ARS to the full MNIST and Cifar-10 data sets, which have $70000$ and $60000$ images, respectively. When using early exaggeration in ARS-BH, we use the ARS version of early exaggeration, given in \eqref{eq:ARS early exaggeration}. Figures \ref{fig:bh_tsne_mnist} and \ref{fig:bh_tsne_cifar} show the results of applying ARS-BH to the MNIST and Cifar-10 data sets, respectively, at $500$, $1000$ and $5000$ iterations with time step $h=1$ using no early exaggeration, $\alpha=1$, and early exaggeration for $250$ iterations with $\alpha=10$. We use the kernel parameters $\theta_1=2$ and $\theta_2=3$, which favor attraction over repulsion, leading to tighter clusters that are better separated. We see in the figures that even after only $500$ iterations the visualization is well-clustered, and the clusters tighten and become more well-separated after more iterations. Even though reasonable results can be obtained without early exaggeration, we do observe improved visualizations when employing early exaggeration.

\begin{figure}[t!]
\centering
\begin{subfigure}{0.32\textwidth}
\includegraphics[width=\textwidth]{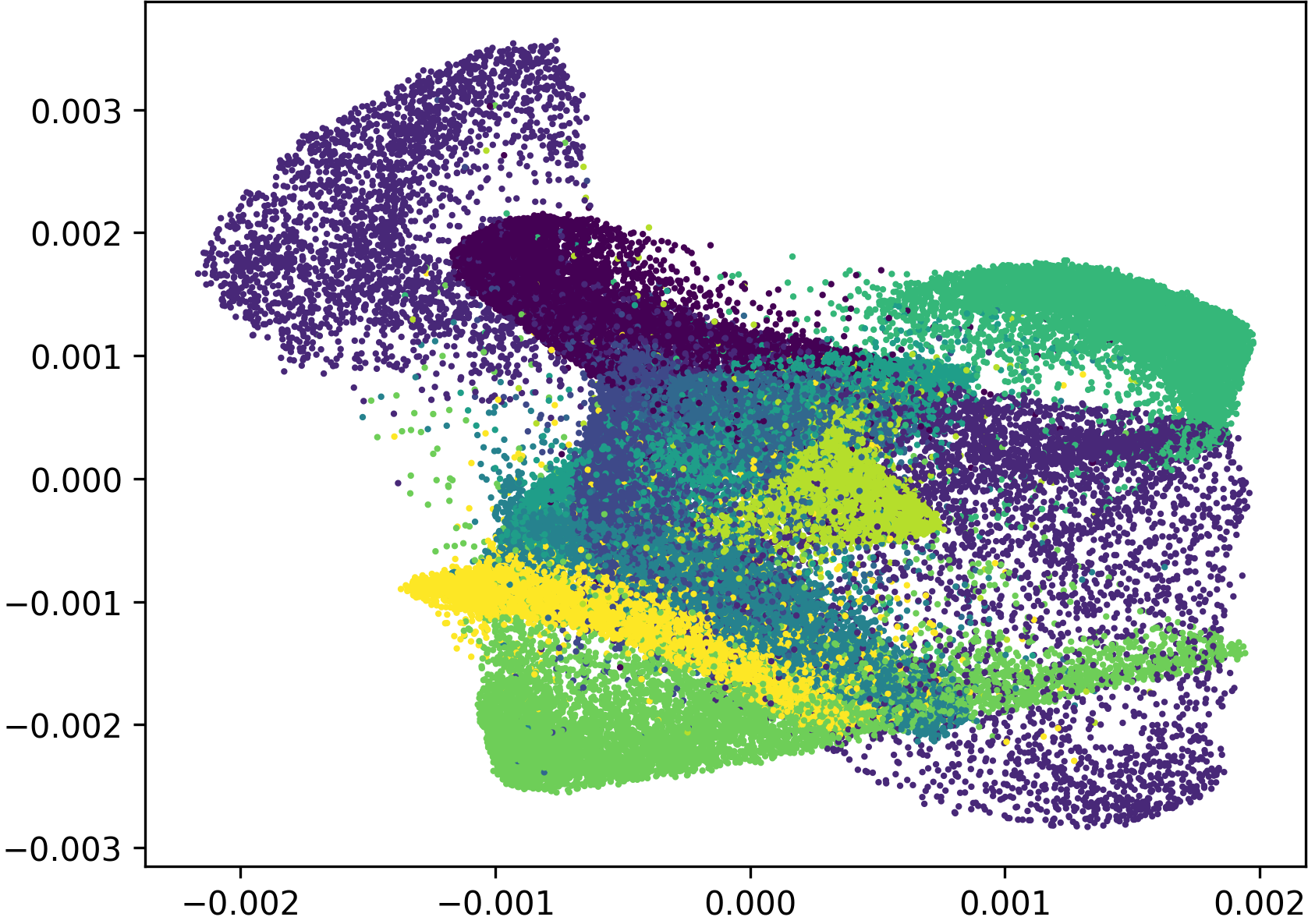}    
\subcaption{MNIST, $T=500$}
\end{subfigure}
\hfill
\begin{subfigure}{0.31\textwidth}
\includegraphics[width=\textwidth]{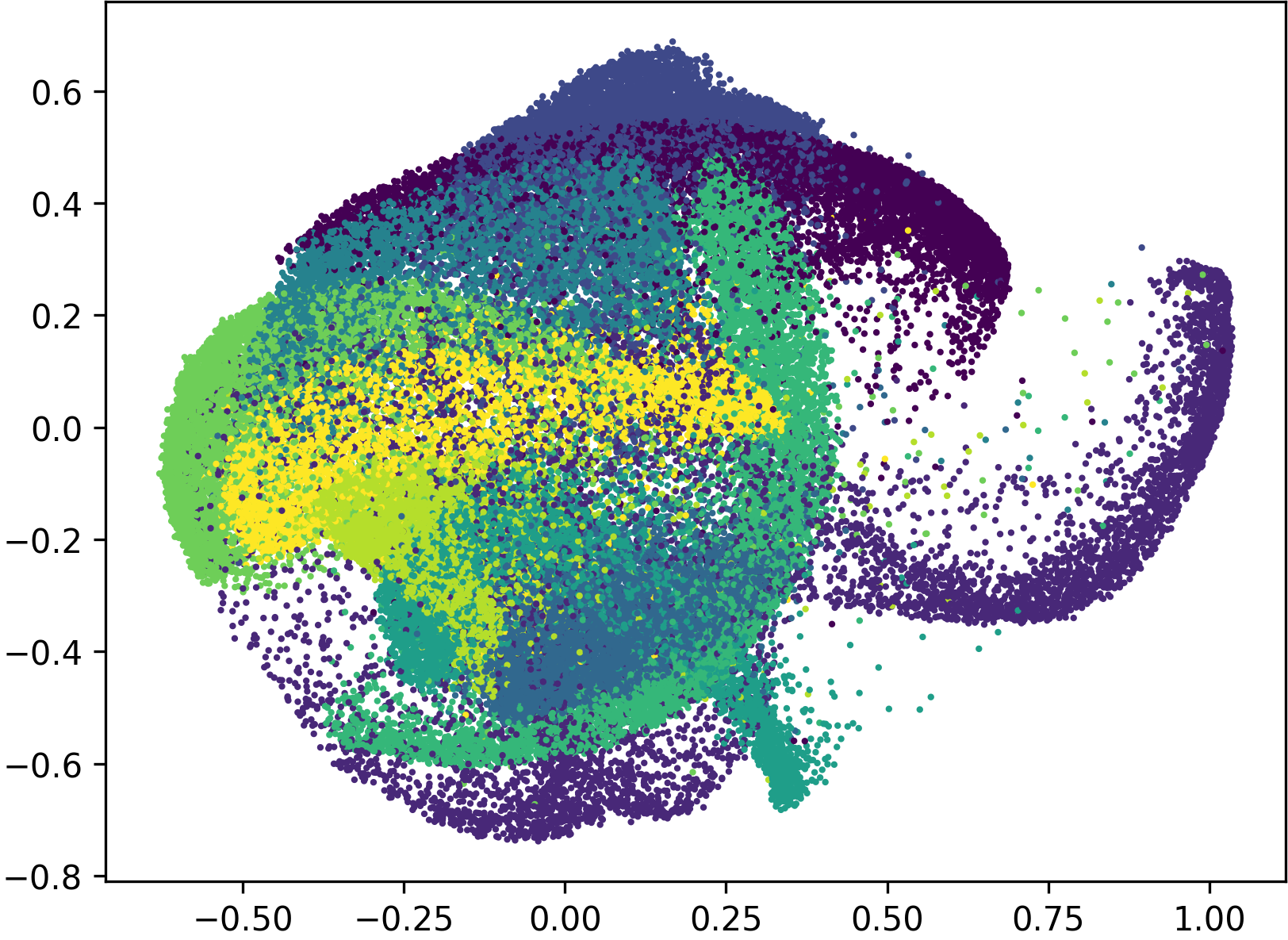}    
\subcaption{MNIST, $T=1000$}
\end{subfigure}
\hfill
\begin{subfigure}{0.305\textwidth}
\includegraphics[width=\textwidth]{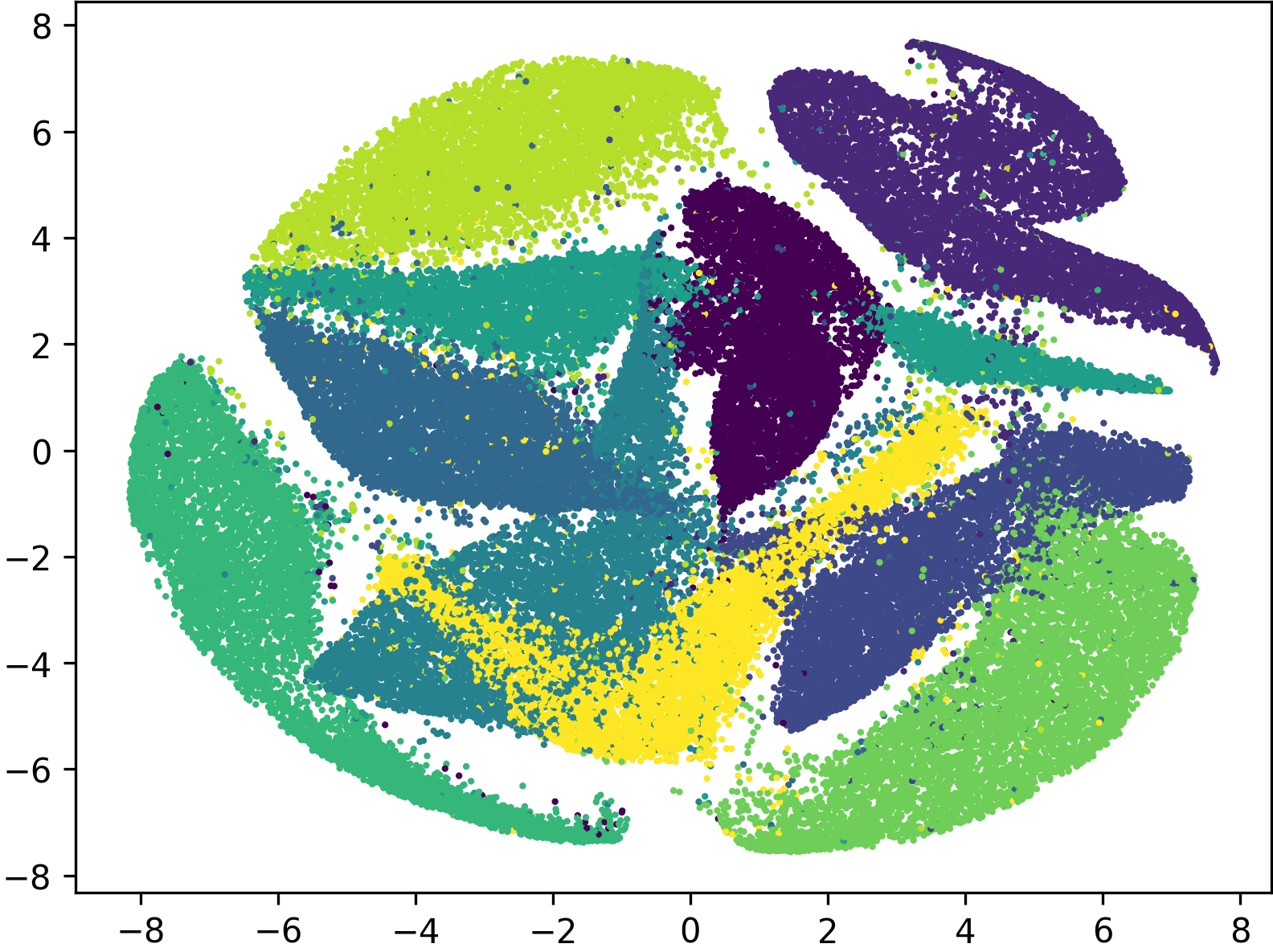}    
\subcaption{MNIST, $T=5000$}
\end{subfigure}\\
\begin{subfigure}{0.32\textwidth}
\includegraphics[width=\textwidth]{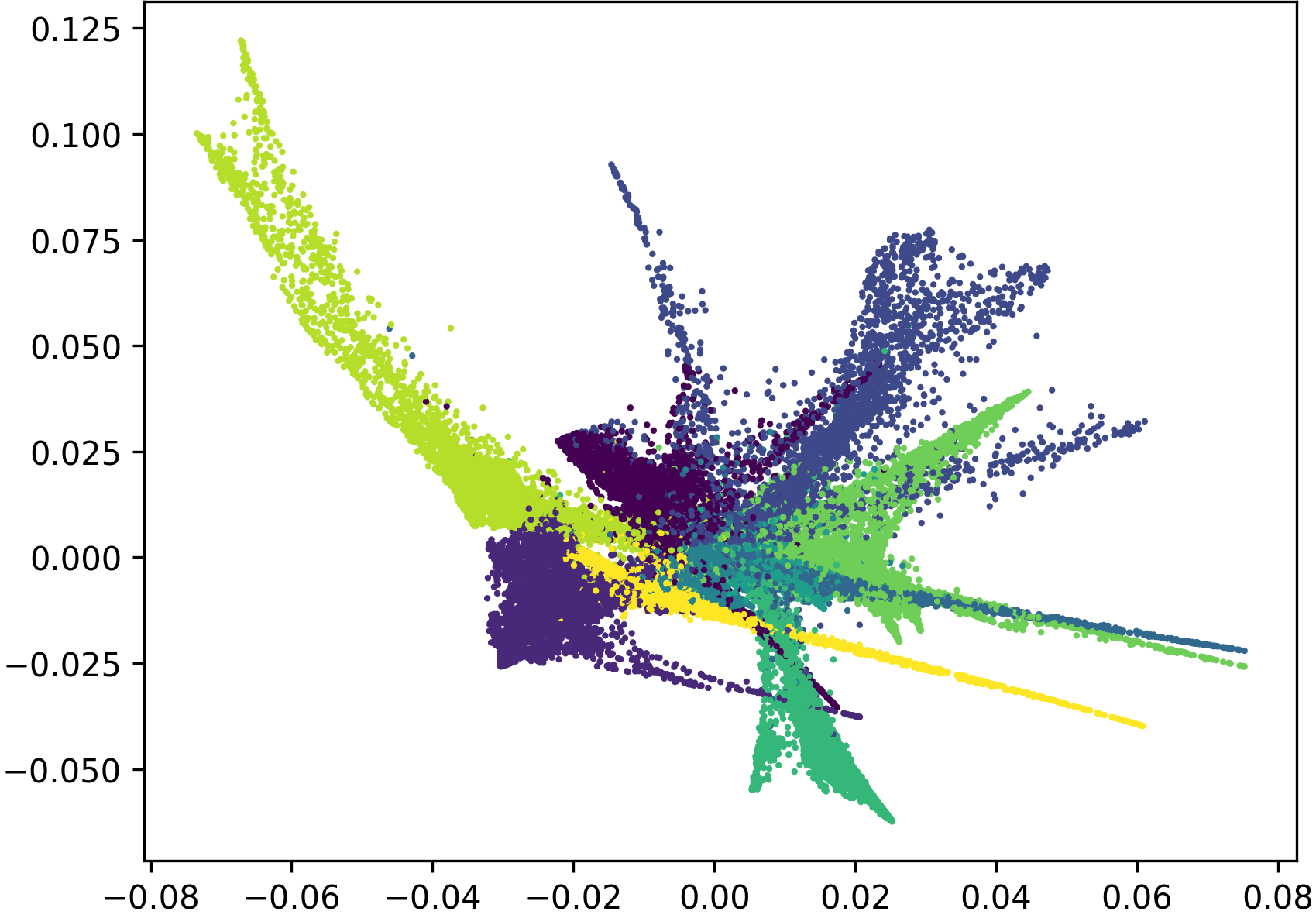}    
\subcaption{Cifar-10, $T=500$}
\end{subfigure}
\hfill
\begin{subfigure}{0.30\textwidth}
\includegraphics[width=\textwidth]{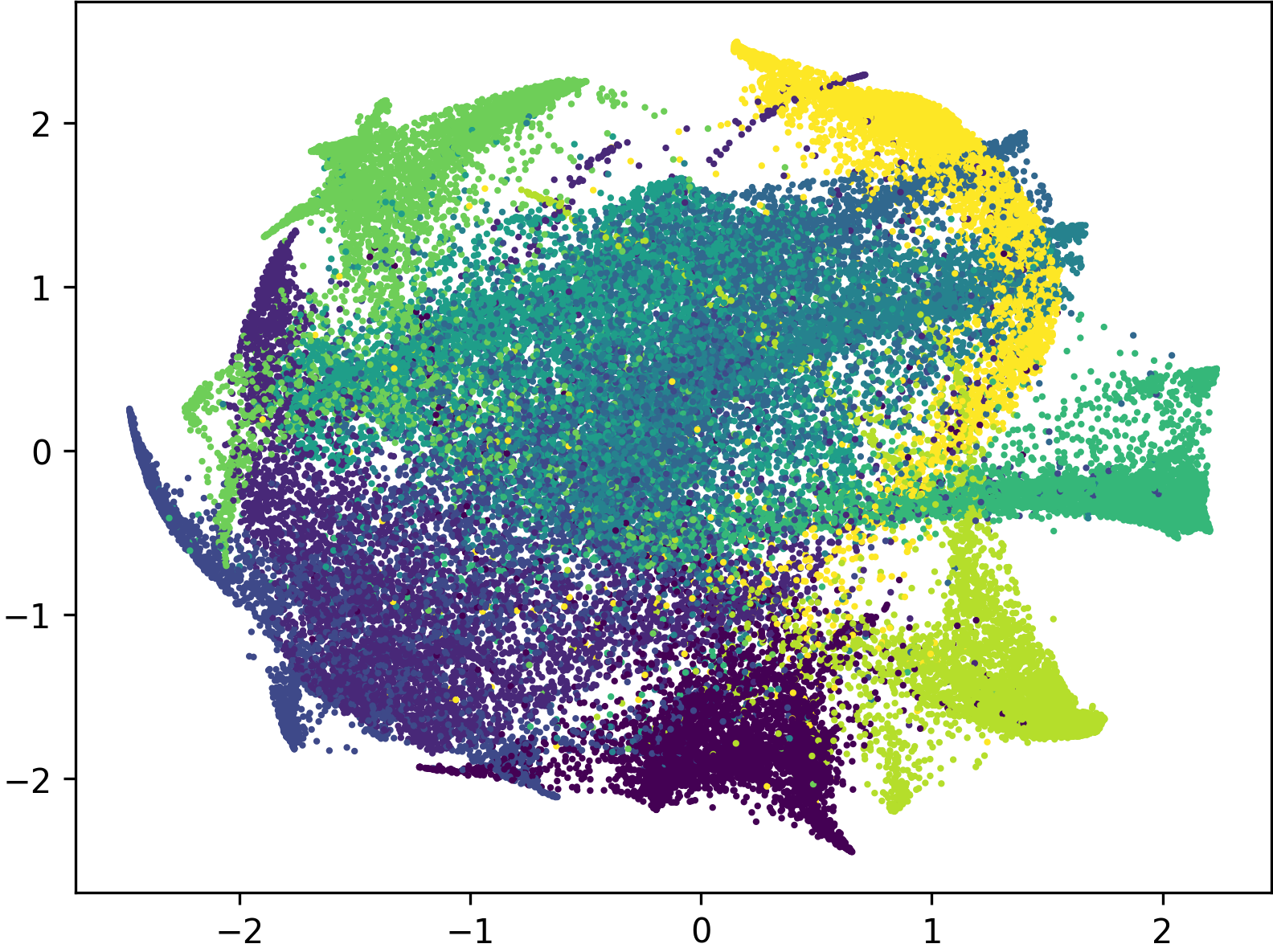}    
\subcaption{Cifar-10, $T=1000$}
\end{subfigure}
\hfill
\begin{subfigure}{0.315\textwidth}
\includegraphics[width=\textwidth]{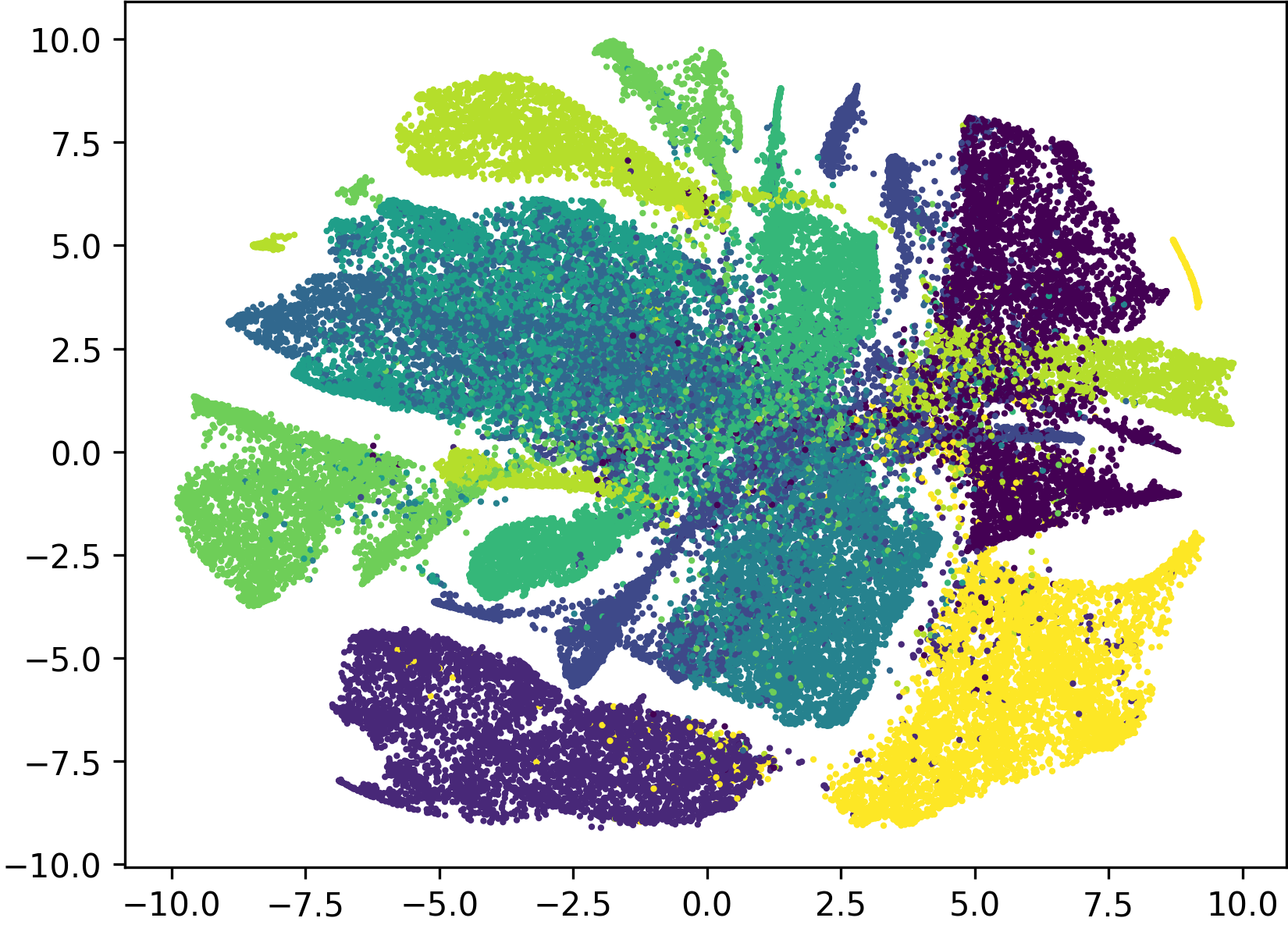}    
\subcaption{Cifar-10, $T=5000$}
\end{subfigure}
\caption{t-SNE with vanilla gradient descent on MNIST and Cifar-10. We used 250 steps of early exaggeration with $\alpha=10$ and a time step of $h=1000$.}
\label{fig:tsne_vanilla}
\end{figure}

It is important to emphasize that ARS-BH is able to obtain good visualizations that are comparable, and in some cases better, than t-SNE, using only the simple iteration \eqref{eq:ARS discrete}, which would be equivalent to running vanilla gradient descent for t-SNE. In particular, we do not need to employ any time-step adaptation or momentum techniques to achieve good results. In contrast, these optimization techniques are crucial for obtaining good results with the t-SNE algorithm. To illustrate this, in Figure \ref{fig:tsne_vanilla} we show the results of running vanilla gradient descent on the t-SNE objective --- the Kullback-Leibler divergence --- after $500$, $1000$, and $5000$ iterations. We used early exaggeration with the same parameters as ARS-BH, which are similar to those used in standard t-SNE implementations.\footnote{For example, in sklearn, the default early exaggeration parameter is $\alpha=12$, while we used $\alpha=10$} We experimented with different choices of the time step $h$, which in vanilla gradient descent is constant in time and not adapted. In Figure \ref{fig:tsne_vanilla} we show the results with $h=1000$, which gave the best results. Smaller time steps $h \ll 1000$ (e.g., $h\in (100,500)$) gave substantially worse results. Even with this fine tuning of the time step and early exaggeration parameters, the results of vanilla gradient descent t-SNE are substantially worse than ARS-BH, as well as the t-SNE visualizations from Figure \ref{fig:ars_tsne}, which employ all the advanced optimization techniques. In contrast, ARS-BH utilized all the same default parameter values as in the earlier experiments in the paper with smaller subsets of MNIST, and does not require any fine tuning. 

\section{Conclusion and Future Work} 

We proposed a new method for data visualization called Attraction-Repulsion Swarming (ARS) visualization. The method is a general framework that can be viewed as a variant of t-SNE, where we abandon the Kullback-Leibler divergence loss function, and instead formulate a dynamical system with attraction-repulsion forces that is run until steady state. By properly normalizing the forces by \emph{total influence}, we obtain dynamics that are far better behaved compared to t-SNE, in that they both converge faster to steady state, and do not require sophisticated optimization tricks to be successful. We furthermore find that favoring attraction forces over repulsion, which is possible in the ARS framework, but not with t-SNE, can in some cases give better visualizations with tighter clusters that are well-separated.

In the future, it would be interesting to perform an in-depth mathematical investigation of the ability of ARS to further explore the tradeoff between attraction and repulsion forces, and to better understand when ARS can uncover clustering structure in data. One possible approach would be to utilize the mean-field limit developed in Section \ref{sec:meanfield}. It would also be interesting to apply the ARS-BH method to a more extensive collection of data sets to evaluate its performance compared to t-SNE.



\bibliography{main}

\begin{thebibliography}{10}

\bibitem{abdelmoula2016data}
W.~M. Abdelmoula, B.~Balluff, S.~Englert, J.~Dijkstra, M.~J. Reinders, A.~Walch, L.~A. McDonnell, and B.~P. Lelieveldt.
\newblock Data-driven identification of prognostic tumor subpopulations using spatially mapped t-sne of mass spectrometry imaging data.
\newblock {\em Proceedings of the National Academy of Sciences}, 113(43):12244--12249, 2016.

\bibitem{barnes1986hierarchical}
J.~Barnes and P.~Hut.
\newblock A hierarchical o (n log n) force-calculation algorithm.
\newblock {\em nature}, 324(6096):446--449, 1986.

\bibitem{belkina2019}
A.~C. Belkina, C.~O. Ciccolella, R.~Anno, R.~Halpert, J.~Spidlen, and J.~E. Snyder-Cappione.
\newblock Automated optimized parameters for t-distributed stochastic neighbor embedding improve visualization and analysis of large datasets.
\newblock {\em Nature communications}, 10(1):5415, 2019.

\bibitem{bocker2022toward}
M.~Bocker, M.~G. Grushko, and K.~E. Arline.
\newblock Toward improved cancer classification using pca+ tsne dimensionality reduction on bulk rna-seq data.
\newblock {\em Cancer Research}, 82(12\_Supplement):2708--2708, 2022.

\bibitem{calder2022graphlearningSOFTWARE}
J.~Calder.
\newblock {GraphLearning Python Package}.
\newblock {\em \emph{\texttt{doi:10.5281/zenodo.5850940}}}, 2022.

\bibitem{chen2020simple}
T.~Chen, S.~Kornblith, M.~Norouzi, and G.~Hinton.
\newblock A simple framework for contrastive learning of visual representations.
\newblock In {\em International conference on machine learning}, pages 1597--1607. PMLR, 2020.

\bibitem{cieslak2020t}
M.~C. Cieslak, A.~M. Castelfranco, V.~Roncalli, P.~H. Lenz, and D.~K. Hartline.
\newblock t-distributed stochastic neighbor embedding (t-sne): A tool for eco-physiological transcriptomic analysis.
\newblock {\em Marine genomics}, 51:100723, 2020.

\bibitem{dolezal2018diagnostic}
J.~M. Dolezal, A.~P. Dash, and E.~V. Prochownik.
\newblock Diagnostic and prognostic implications of ribosomal protein transcript expression patterns in human cancers.
\newblock {\em BMC cancer}, 18:1--14, 2018.

\bibitem{gang2018dimensionality}
P.~Gang, W.~Zhen, W.~Zeng, Y.~Gordienko, Y.~Kochura, O.~Alienin, O.~Rokovyi, and S.~Stirenko.
\newblock Dimensionality reduction in deep learning for chest x-ray analysis of lung cancer.
\newblock In {\em 2018 tenth international conference on advanced computational intelligence (ICACI)}, pages 878--883. IEEE, 2018.

\bibitem{gray2000n}
A.~Gray and A.~Moore.
\newblock N-body'problems in statistical learning.
\newblock {\em Advances in neural information processing systems}, 13, 2000.

\bibitem{hamid2020t}
Y.~Hamid and M.~Sugumaran.
\newblock A t-sne based non linear dimension reduction for network intrusion detection.
\newblock {\em International Journal of Information Technology}, 12:125--134, 2020.

\bibitem{HT2017}
S.~He and E.~Tadmor.
\newblock Global regularity of two-dimensional flocking hydrodynamics.
\newblock {\em Comptes Rendus Mathematique}, 355(7):795--805, 2017.

\bibitem{HT2020}
S.~He and E.~Tadmor.
\newblock A game of alignment: collective behavior of multi-species.
\newblock In {\em Annales de l'Institut Henri Poincar{\'e} C, Analyse non lin{\'e}aire}, volume~38, pages 1031--1053. Elsevier, 2021.

\bibitem{jacobs1988increased}
R.~A. Jacobs.
\newblock Increased rates of convergence through learning rate adaptation.
\newblock {\em Neural networks}, 1(4):295--307, 1988.

\bibitem{kobak2019}
D.~Kobak and P.~Berens.
\newblock The art of using t-sne for single-cell transcriptomics.
\newblock {\em Nature communications}, 10(1):5416, 2019.

\bibitem{kobak2019heavy}
D.~Kobak, G.~Linderman, S.~Steinerberger, Y.~Kluger, and P.~Berens.
\newblock Heavy-tailed kernels reveal a finer cluster structure in t-sne visualisations.
\newblock In {\em Joint European Conference on Machine Learning and Knowledge Discovery in Databases}, pages 124--139. Springer, 2019.

\bibitem{krizhevsky2009learning}
A.~Krizhevsky, G.~Hinton, et~al.
\newblock Learning multiple layers of features from tiny images.
\newblock {\em University of Toronto}, 2009.

\bibitem{lecun1998gradient}
Y.~LeCun, L.~Bottou, Y.~Bengio, and P.~Haffner.
\newblock Gradient-based learning applied to document recognition.
\newblock {\em Proceedings of the IEEE}, 86(11):2278--2324, 1998.

\bibitem{li2017application}
W.~Li, J.~E. Cerise, Y.~Yang, and H.~Han.
\newblock Application of t-sne to human genetic data.
\newblock {\em Journal of bioinformatics and computational biology}, 15(04):1750017, 2017.

\bibitem{linderman2019fast}
G.~C. Linderman, M.~Rachh, J.~G. Hoskins, S.~Steinerberger, and Y.~Kluger.
\newblock Fast interpolation-based t-sne for improved visualization of single-cell rna-seq data.
\newblock {\em Nature methods}, 16(3):243--245, 2019.

\bibitem{linderman2019clustering}
G.~C. Linderman and S.~Steinerberger.
\newblock Clustering with t-sne, provably.
\newblock {\em SIAM journal on mathematics of data science}, 1(2):313--332, 2019.

\bibitem{linderman2022dimensionality}
G.~C. Linderman and S.~Steinerberger.
\newblock Dimensionality reduction via dynamical systems: the case of t-sne.
\newblock {\em SIAM Review}, 64(1):153--178, 2022.

\bibitem{mandel2020sequential}
J.~Mandel, R.~Avula, and E.~V. Prochownik.
\newblock Sequential analysis of transcript expression patterns improves survival prediction in multiple cancers.
\newblock {\em BMC cancer}, 20:1--14, 2020.

\bibitem{mandel2019expression}
J.~Mandel, H.~Wang, D.~P. Normolle, W.~Chen, Q.~Yan, P.~C. Lucas, P.~V. Benos, and E.~V. Prochownik.
\newblock Expression patterns of small numbers of transcripts from functionally-related pathways predict survival in multiple cancers.
\newblock {\em BMC cancer}, 19(1):1--15, 2019.

\bibitem{miyamoto2022natural}
A.~Miyamoto, M.~V. Bendarkar, and D.~N. Mavris.
\newblock Natural language processing of aviation safety reports to identify inefficient operational patterns.
\newblock {\em Aerospace}, 9(8):450, 2022.

\bibitem{motsch2011}
S.~Motsch and E.~Tadmor.
\newblock A new model for self-organized dynamics and its flocking behavior.
\newblock {\em Journal of Statistical Physics}, 144:923--947, 2011.

\bibitem{murray2024large}
R.~Murray and A.~Pickarski.
\newblock Large data limits and scaling laws for tsne.
\newblock {\em arXiv preprint arXiv:2410.13063}, 2024.

\bibitem{olszewski2024dimensionality}
D.~Olszewski, M.~Iwanowski, and W.~Graniszewski.
\newblock Dimensionality reduction for detection of anomalies in the iot traffic data.
\newblock {\em Future Generation Computer Systems}, 151:137--151, 2024.

\bibitem{platzer2013visualization}
A.~Platzer.
\newblock Visualization of snps with t-sne.
\newblock {\em PloS one}, 8(2):e56883, 2013.

\bibitem{sanders2021unmasking}
A.~C. Sanders, R.~C. White, L.~S. Severson, R.~Ma, R.~McQueen, H.~C.~A. Paulo, Y.~Zhang, J.~S. Erickson, and K.~P. Bennett.
\newblock Unmasking the conversation on masks: Natural language processing for topical sentiment analysis of covid-19 twitter discourse.
\newblock {\em AMIA Summits on Translational Science Proceedings}, 2021:555, 2021.

\bibitem{sharma2021stochastic}
N.~Sharma, M.~Mangla, S.~N. Mohanty, and S.~Satpaty.
\newblock A stochastic neighbor embedding approach for cancer prediction.
\newblock In {\em 2021 international conference on emerging smart computing and informatics (ESCI)}, pages 599--603. IEEE, 2021.

\bibitem{sharma2023optimization}
N.~Sharma and S.~Sharma.
\newblock Optimization of t-sne by tuning perplexity for dimensionality reduction in nlp.
\newblock In {\em International Conference on Communication and Computational Technologies}, pages 519--528. Springer, 2023.

\bibitem{shetty2021automated}
P.~Shetty and R.~Ramprasad.
\newblock Automated knowledge extraction from polymer literature using natural language processing.
\newblock {\em Iscience}, 24(1), 2021.

\bibitem{steinerberger2022t}
S.~Steinerberger and Y.~Zhang.
\newblock t-sne, forceful colorings, and mean field limits.
\newblock {\em Research in the Mathematical Sciences}, 9(3):42, 2022.

\bibitem{taskesen20162d}
E.~Taskesen and M.~J. Reinders.
\newblock 2d representation of transcriptomes by t-sne exposes relatedness between human tissues.
\newblock {\em PloS one}, 11(2):e0149853, 2016.

\bibitem{van2014accelerating}
L.~Van Der~Maaten.
\newblock Accelerating t-sne using tree-based algorithms.
\newblock {\em The journal of machine learning research}, 15(1):3221--3245, 2014.

\bibitem{van2008visualizing}
L.~Van~der Maaten and G.~Hinton.
\newblock Visualizing data using t-sne.
\newblock {\em Journal of Machine Learning Research}, 9(11), 2008.

\bibitem{wang2018comparison}
Y.~Wang, S.~Liu, N.~Afzal, M.~Rastegar-Mojarad, L.~Wang, F.~Shen, P.~Kingsbury, and H.~Liu.
\newblock A comparison of word embeddings for the biomedical natural language processing.
\newblock {\em Journal of biomedical informatics}, 87:12--20, 2018.

\bibitem{xue2020classification}
J.~Xue, Y.~Chen, O.~Li, and F.~Li.
\newblock Classification and identification of unknown network protocols based on cnn and t-sne.
\newblock In {\em Journal of Physics: Conference Series}, volume 1617, page 012071. IOP Publishing, 2020.

\bibitem{yang2009heavy}
Z.~Yang, I.~King, Z.~Xu, and E.~Oja.
\newblock Heavy-tailed symmetric stochastic neighbor embedding.
\newblock {\em Advances in neural information processing systems}, 22, 2009.

\bibitem{yi2021anomaly}
H.~Yi-ran, S.~Yi-qiang, and W.~Jing-lin.
\newblock Anomaly detection of network traffic based on t-sne dimensionality reduction preprocessing.
\newblock {\em Computer and Modernization}, (02):109, 2021.

\bibitem{zhou2022classification}
L.~Zhou, M.~Rueda, and A.~Alkhateeb.
\newblock Classification of breast cancer nottingham prognostic index using high-dimensional embedding and residual neural network.
\newblock {\em Cancers}, 14(4):934, 2022.

\bibitem{zoghi2021unsw}
Z.~Zoghi and G.~Serpen.
\newblock Unsw-nb15 computer security dataset: Analysis through visualization.
\newblock {\em arXiv preprint arXiv:2101.05067}, 2021.

\end{thebibliography}
\bibliographystyle{abbrv}

\appendix

\section{Derivation of Mean-Field Limit}\label{appendix: mean-field}

We derive the mean-field equations \eqref{eq:ARS mean-field} from the particle system \eqref{eq: ARS}. We test $\partial_t \rho_t$ against arbitrary smooth function $\varphi$,
\begin{equation}\label{eq:rho test}
\begin{split}
\int \partial_t\rho_t(\by)\varphi(\by)d\by &= \frac{d}{dt}[\frac{1}{N}\sum^N_{i = 1}\varphi(\by_i(t))]\\
& = \frac{1}{N}\sum^N_{i=1} \dot{\by}_i\cdot\nabla\varphi(\by_i) \\
& = \frac{1}{N}\sum^N_{i=1} \bu_i\cdot\nabla\varphi(\by_i).
\end{split}
\end{equation}
The velocity $\bu_i$ reflects the competition between attraction and repulsion forcing, 
\[
\begin{split}
\bu_i &= \frac{1}{\sum^N_{j=1} P_{ij}}\sum^N_{j=1}P_{ij}\psi_1(|\by_i-\by_j|)(\by_j-\by_i)-\frac{1}{\sum^N_{j=1} Q_{ij}}\sum^N_{j=1}Q_{ij}\psi_2(|\by_i-\by_j|)(\by_j-\by_i)\\
& = \frac{1}{N}\sum^N_{j=1}[\frac{P_{ij}}{\bar{P}_i}\psi_1(|\by_i-\by_j|)-\frac{Q_{ij}}{\bar{Q}_i}\psi_2(|\by_i-\by_j|)](\by_j-\by_i),
\end{split}
\]
where 
\[
\bar{P}_i = \frac{1}{N}\sum^N_{j=1}P_{ij}, \quad \bar{Q}_i = \frac{1}{N}\sum^N_{j=1}Q_{ij}.
\]

The normalized similarity, $\displaystyle\frac{P_{ij}}{\bar{P}_i}$, can be written as
\[
\begin{split}
\frac{P_{ij}}{\bar{P}_i} &= \frac{e^{-\frac{|\bx_i-\bx_j|^2}{2\sig^2}}}{\frac{1}{N}\sum^N_{k=1}e^{-\frac{|\bx_i-\bx_k|^2}{2\sig^2}}}\\
&= \frac{e^{-\frac{|(T^N_t)^{-1}\by_i-(T^N_t)^{-1}\by_j|^2}{2\sig^2}}}{\frac{1}{N}\sum^N_{k=1}e^{-\frac{|(T^N_t)^{-1}\by_i-(T^N_t)^{-1}\by_k|^2}{2\sig^2}}}\\
& = \frac{e^{-\frac{|(T^N_t)^{-1}\by_i-(T^N_t)^{-1}\by_j|^2}{2\sig^2}}}{\int_{\R^d}e^{-\frac{|(T^N_t)^{-1}\by_i-(T^N_t)^{-1}\bz|^2}{2\sig^2}}\rho_t(\bz)d\bz}
\end{split}
\]
where the projection $T^N_t: \R^D\mapsto \R^d$ defines the one-to-one mapping between the input data and the output embedding, $T^N_t\bx_i = \by_i$, $1\leq i\leq N$, in the $N-$particle system \eqref{eq: ARS}. By formally passing the mean-field limit $N\rightarrow\infty$, the pre-image of $\displaystyle T_t = \lim_{N\rightarrow\infty}T^N_t$ gives a level set
\[
S_t(\by) := T^{-1}_t\by = \{\bx\in\supp\{\mu\}: T_t\bx = \by\}.
\]
Hence, we have
\begin{equation}\label{eq:P mean-field}
\tP(\by_i,\by_j,t) := \lim_{N\rightarrow\infty}\frac{P_{ij}}{\bar{P}_i} = \frac{\iint_{S_t(\by)\times S_t(\by')} e^{-\frac{|\bx-\bx'|}{2\sig^2}}\mu(\bx)\mu(\bx')d\bx d\bx'}{\int_{\R^d}[\iint_{S_t(\by)\times S_t(\bz)} e^{-\frac{|\bx-\bx'|}{2\sig^2}}\mu(\bx)\mu(\bx')d\bx d\bx']\rho_t(\bz)d\bz}. 
\end{equation}
The mean-field limit for $\displaystyle\frac{Q_{ij}}{\bar{Q}_i}$ is obtained with
\begin{equation}\label{eq:Q mean-field}
\begin{split}
\tQ(\by_i,\by_j,t) := \lim_{N\rightarrow\infty} \frac{Q_{ij}}{\bar{Q}_i}  &= \lim_{N\rightarrow\infty} \frac{\frac{1}{1+|\by_i-\by_j|^2}}{\frac{1}{N}\sum^N_{k=1}\frac{1}{1+|\by_i-\by_k|^2}} = \frac{1}{\int\frac{1}{1+|\by_i-\bz|^2}\rho_t(\bz)d\bz}\cdot\frac{1}{1+|\by_i-\by_j|^2}
\end{split}
\end{equation}
Then \eqref{eq:P mean-field}, \eqref{eq:Q mean-field} yield 
\begin{equation} \label{eq:u mean-field}
    \bu(\by_i,t) := \lim_{N\rightarrow\infty}\bu_i = \int_{R^d}[\tP(\by_i,\by',t)\psi_1(|\by_i-\by'|)-\tQ(\by_i,\by',t)\psi_2(|\by_i-\by'|)](\by'-\by_i)\rho_t(\by')d\by'
\end{equation}
Subsequently, formal integration by parts in \eqref{eq:rho test} yields
\[
\begin{split}
\int_{\R^d} \partial_t\rho_t(\by)\varphi(\by) d\by & = \int_{\R^d}\bu(\by,t)\cdot\nabla\varphi(\by)\rho_t(\by)d\by\\
& = -\int_{\R^d}\nabla\cdot(\bu(\by,t)\rho_t(\by))\varphi(\by)d\by.
\end{split}
\]
Since the test function $\varphi$ is arbitrary, we obtain the mean-field equation 
\[
\partial_t\rho_t+\nabla\cdot(\rho_t\bu) = 0.
\]

\end{document}